\documentclass[sn-mathphys-num]{sn-jnl}

\usepackage{graphicx}%
\usepackage{multirow}%
\usepackage{amsmath,amssymb,amsfonts}%
\usepackage{amsthm}%
\usepackage{mathrsfs}%
\usepackage[title]{appendix}%
\usepackage{xcolor}%
\usepackage{textcomp}%
\usepackage{manyfoot}%
\usepackage{booktabs}%
\usepackage{algorithm}%
\usepackage{algorithmicx}%
\usepackage{algpseudocode}%
\usepackage{listings}%
\usepackage{graphicx}
\usepackage{amsmath}
\usepackage{xcolor,colortbl}
\usepackage{color, colortbl}
\definecolor{tabhighlight}{HTML}{e5e5e5}
\definecolor{citecolor}{HTML}{0071bc}
\newcommand{\tableCellHeight}{1}
\newcommand{\tabstyle}[1]{
  \setlength{\tabcolsep}{#1}
  \renewcommand{\arraystretch}{\tableCellHeight}
  \centering
  \small
}
\usepackage{subcaption}
\usepackage{sidecap}
\newcommand{\tablestyle}[2]{\setlength{\tabcolsep}{#1}\renewcommand{\arraystretch}{#2}\centering\footnotesize}
\usepackage{color}
\def\blu#1{\textbf{\color{blue} #1}} 
\def\red#1{\textbf{\color{red}  #1}} 
\usepackage{newfloat}
\usepackage{listings}
\usepackage[capitalize]{cleveref}

\usepackage{multicol}

\theoremstyle{thmstyleone}%
\theoremstyle{thmstyletwo}%

\theoremstyle{thmstylethree}%

\begin{document}

\title[Promoting Equity: Generalized Domain Prompt Learning for Accessible VLM Research]{Promoting AI Equity in Science: Generalized Domain Prompt Learning for Accessible VLM Research}

\author[1,2]{\fnm{Qinglong} \sur{Cao}}

\author*[2]{\fnm{Yuntian} \sur{Chen}}\email{ychen@eitech.edu.cn}

\author[3]{\fnm{Lu} \sur{Lu}}

\author[5]{\fnm{Hao} \sur{Sun}}

\author[4]{\fnm{Zhenzhong} \sur{Zeng}}


\author[1]{\fnm{Xiaokang} \sur{Yang}}


\author[2]{\fnm{Dongxiao} \sur{Zhang}}

\affil[1]{\orgdiv{MoE Key Lab of Artificial Intelligence, AI Institute}, \orgname{Shanghai Jiao Tong University,}, \orgaddress{Shanghai 200240, China}}

\affil[2]{Ningbo Institute of Digital Twin, Eastern Institute of Technology,
Ningbo 315200, China}

\affil[3]{Department of Statistics and Data Science, Yale University, New Haven, CT 06511, USA}

\affil[4]{School of Environmental Science and Engineering, Southern University of Science and Technology, Shenzhen, China}

\affil[5]{Gaoling School of Artificial Intelligence, Renmin University of China, Beijing, China}



\abstract{
Large-scale Vision-Language Models (VLMs) have demonstrated exceptional performance in natural vision tasks, motivating researchers across domains to explore domain-specific VLMs. However, the construction of powerful domain-specific VLMs demands vast amounts of annotated data, substantial electrical energy, and computing resources, primarily accessible to industry, yet hindering VLM research in academia. To address this challenge and foster sustainable and equitable VLM research, we present the Generalized Domain Prompt Learning (GDPL) framework. GDPL facilitates the transfer of VLMs' robust recognition capabilities from natural vision to specialized domains, without the need for extensive data or resources. By leveraging small-scale domain-specific foundation models and minimal prompt samples, GDPL empowers the language branch with domain knowledge through quaternion networks, uncovering cross-modal relationships between domain-specific vision features and natural vision-based contextual embeddings. Simultaneously, GDPL guides the vision branch into specific domains through hierarchical propagation of generated vision prompt features, grounded in well-matched vision-language relations. Furthermore, to fully harness the domain adaptation potential of VLMs, we introduce a novel low-rank adaptation approach. Extensive experiments across diverse domains like remote sensing, medical imaging, geology, Synthetic Aperture Radar, and fluid dynamics, validate the efficacy of GDPL, demonstrating its ability to achieve state-of-the-art domain recognition performance in a prompt learning paradigm. Our framework paves the way for sustainable and inclusive VLM research, transcending the barriers between academia and industry.

}

\maketitle

\section{Introduction}\label{sec1}

\begin{figure*}[ht]
	\begin{center}
		\includegraphics[width=1.0\linewidth]{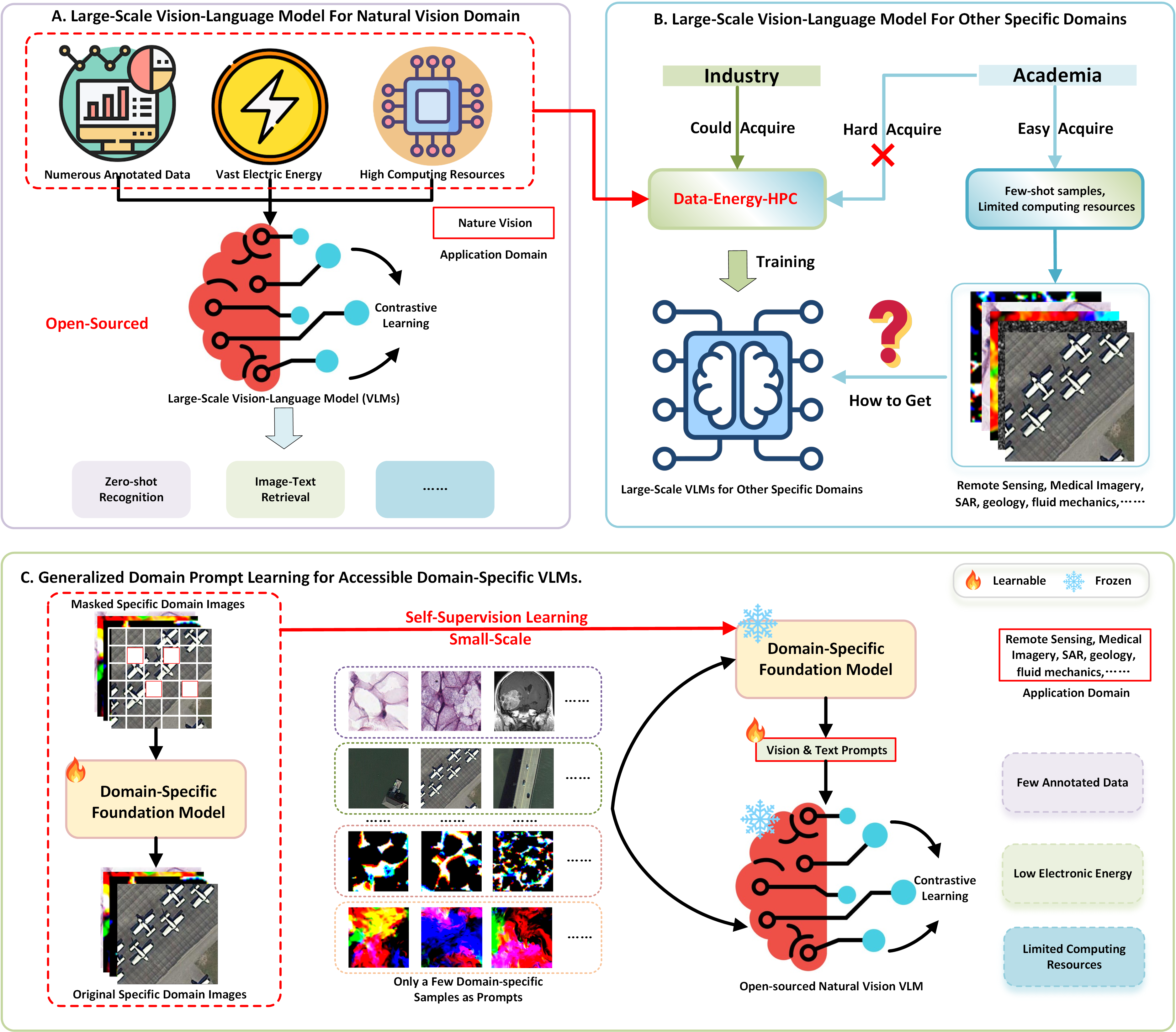}
	\end{center}
	\caption{The main concept of the proposed method. \textbf{(A)} Existing Large-scale vision-language models (VLMs) for the natural vision domain. \textbf{(B)} The same logic of \textbf{(A)} to acquire VLMs for specific domains is possible for the industry yet inaccessible for academia. \textbf{(C)} Domain-specific foundation models prompt VLMs into specific domains. HPC: High Performance Computing Center. }
	\label{overall}
\end{figure*}

Fully-supervised deep learning methods with large-scale training data have achieved remarkable success in various visual or language understanding tasks~\cite{simonyan2014very,he2016deep,long2015fully,chen2017deeplab,redmon2016you}. Particularly, with the emergence of large-scale Vision-Language Models (VLMs)~\cite{radford2021learning,su2019vl,jia2021scaling,zhang2021vinvl}, the vision understanding task has achieved a new milestone. Existing VLMs concentrate on establishing robust cross-modal relationships via contrastive learning among aligned vision-language pairs. Among these models, Contrastive Language-Image Pre-training (CLIP)~\cite{radford2021learning} is widely acclaimed for its remarkable generalization capabilities and extensive training of vision-language pairs. The success of VLMs in natural vision domain has inspired many researchers across other science domains to explore the domain-specific VLMs~\cite{chen2024towards,lu2024visual}. However, as shown in Figure~\ref{overall}, developing robust and powerful domain-specific VLMs necessitates extensive collections of annotated data, substantial electrical energy, and computational resources, which are accessible to industry. This situation poses a significant obstacle to VLM research in academia.

To address this issue, and advance fair VLM research for academia, we argue that only a centrical large-scale VLM is needed, the centrical VLM could be transferred to the targeted domain VLMs with small-scale domain-specific foundation models in a generalized domain prompt learning manner. Prompt learning aims to adapt VLMs to downstream tasks using learned prompts from limited training data, thus potentially addressing the issue of insufficient real-world data in practical scenarios~\cite{zhou2022learning, zhou2022conditional, khattak2023maple, wang2022learning, luddecke2022image, rao2022denseclip,ge2023domain}. Particularly, when applying CLIP for zero-shot recognition for downstream tasks, how to manually give appropriate prompts becomes challenging. To handle this problem, CoOp~\cite{zhou2022learning}, the first prompt learning approach for VLMs, creatively introduced context optimization to provide learnable language prompts, which effectively boosts the adaptability of CLIP for downstream recognition tasks.  Yet, the plain context optimization would inevitably cause class-shifting issues and compromise the performance due to the overfitting. Correspondingly, CoCoOp\cite{zhou2022conditional}  proposed input-conditional language prompts for the language branch to refine the vision-language alignment and alleviate the class-shifting. Previous methods mainly focus on providing language prompts as inputs to acquire better performance for downstream tasks. Differently, MaPLe~\cite{khattak2023maple} constructs a cross-modal prompt learning paradigm and simultaneously provides prompts for both vision and language branches, which results in more stable vision-language alignment with better performance.

While recent advances in prompt learning have showcased impressive results, their focus has primarily been on natural vision tasks. The application of VLMs in other specialized domains like remote sensing, medical imagery, geology, Synthetic Aperture Radar (SAR), and fluid dynamics remains a significant challenge, and the inherent lack of domain-specific adaptation limits their effectiveness in practical vertical domains. Prompting VLMs with domain-specific inputs often results in misinterpretation within the context of natural vision, leading to considerable performance degradation. This limitation impedes the widespread utilization of large-scale models in critical real-world sectors. To address this pressing concern and inspired by the utilization of domain knowledge~\cite{xu2024worth}, we propose a novel framework: Generalized Domain Prompt Learning. Our framework aims to facilitate the seamless transfer of VLMs' robust recognition capabilities from constrained natural vision settings to specialized domains, leveraging the domain knowledge from the domain-specific foundation models. By doing so, we aim to bridge the gap between the capabilities of current large models and the requirements of real-world applications in diverse vertical domains, extending beyond traditional natural language and image processing tasks.  The overall pipeline of our proposed method is shown in Figure~\ref{overall}. Particularly, existing VLMs leverage the contrastive pre-training and large-scale vision-text pairs to establish well-matched vision-language relationships for natural images and have shown powerful zero-shot recognition performance. To prompt the powerful VLMs into the specific domains, we could find the open-sourced domain-specific foundation models or directly leverage relatively limited datasets of specific domains to obtain the domain-specific foundation models in a masked autoencoder (MAE)~\cite{he2022masked} pattern.  Utilizing the domain-specific foundation models to provide the domain knowledge, we respectively prompt the vision and language branches into the specific domain and achieve great zero-shot recognition performance on specific domains such as remote sensing, medical imagery, geology, SAR, and fluid dynamics.

It is worth noting that the majority of existing open-source domain-specific foundation models~\cite{wang2022advancing,sun2022ringmo,ma2023segment} are primarily pre-trained solely at the vision level, lacking inherent support for simultaneous prompting of both vision and language modalities. Inspired by the quaternion network~\cite{parcollet2018quaternion}, renowned for its adeptness in exploring inter- and intra-correlations within the quaternion hidden space, domain-specific vision features from the domain-specific foundation model and generalized contextual embeddings from the language branch are initially fed into quaternion networks to uncover cross-modal relationships. This process results in the generation of domain-specific contextual embeddings for the language branch. Meanwhile, pre-trained VLMs have established a robust and consistent vision-language matching relationship. Consequently, domain knowledge can be readily propagated from the specialized language branch of VLMs to the vision branch. Specifically, we utilize trainable language prompt features as input and domain-specific vision features as guidance to probe the intermodal relationships within each vision-language layer. This hierarchical process could yield vision prompt features for each visual layer and progressively transfer the vision branch into specialized space.

Furthermore, for optimal exploitation of the domain adaptation capabilities inherent in VLMs, we may consider employing Low-Rank Adaptation (LoRA) techniques ~\cite{hu2021lora}, which was Originally introduced as a fine-tuning method, which entails the utilization of trainable low-rank layers to update the original features with a minimal number of parameters. However, applying LoRA in its original form, without considering the modalities involved, risks disrupting the established stability of the vision-language matching relationship. To address this challenge, we introduce a novel approach wherein a learnable cross-modal matrix is employed to govern the update process of low-rank parameters. Specifically, with independent low-rank parameters for both the vision and language branches, the cross-modal matrix facilitates the fusion of these parameters, enabling the computation of modal-shift parameters. These modal-shift parameters are subsequently added to the original low-rank parameters to ensure consistent updates across modalities. Through this innovative mechanism, we effectively mine the domain adaptation potential while preserving the integrity of the matched vision-language relationship.

To demonstrate the effectiveness of our proposed method, we conducted experiments across various specific domains, including remote sensing, medical imaging,  geology, SAR, and fluid dynamics. All experimental outcomes unequivocally demonstrate that the proposed method significantly outperforms previous natural vision-based methods.

	



To sum up, the proposed method contributes in the following ways:
\begin{itemize}
	\item Introducing Generalized Domain Prompt Learning (GDPL), a novel framework that harnesses domain-specific foundation models and minimal prompts to enrich VLMs with domain knowledge. This approach empowers VLMs to adapt their recognition capabilities to specialized domains, fostering accessible research across various scientific fields.

	\item Proposing a novel low-rank adaptation approach within GDPL, which optimizes VLMs' domain adaptation potential while preserving the integrity of the vision-language relationship. This enhancement ensures the adaptability and robustness of VLMs across diverse specialized domains.

	\item Successfully addressing the challenge of domain-specific adaptation in VLMs through GDPL, thereby opening up new avenues for VLM applications. Extensive experimentation across diverse domain-specific datasets validates the efficacy of GDPL, demonstrating significant advancements in domain transfer tasks and contributing to the promotion of equity in VLM research.
\end{itemize}

 \begin{figure*}[ht]
	\begin{center}
		\includegraphics[width=1.0\linewidth]{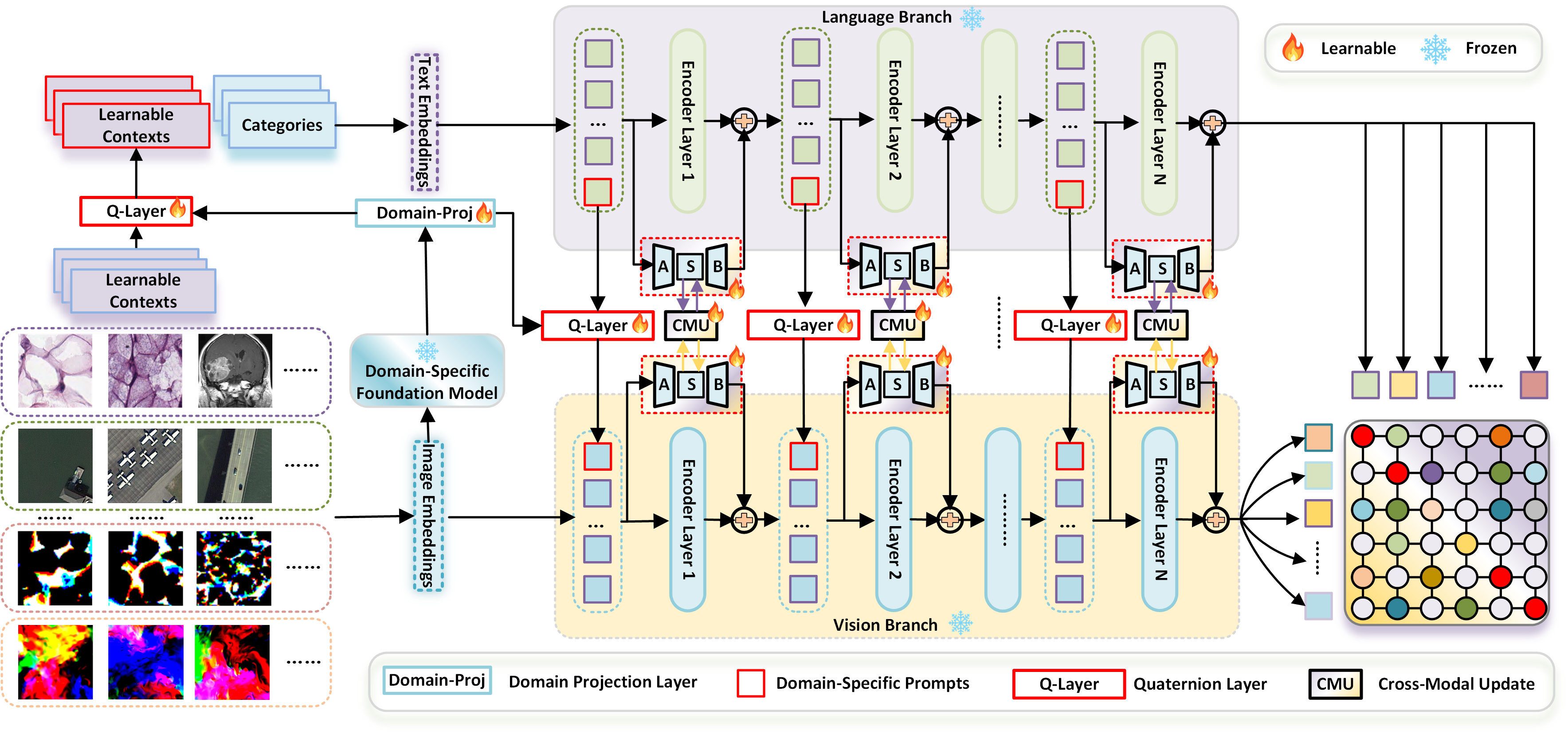}
	\end{center}
	\caption{The network of our proposed generalized domain prompt learning. Utilizing the domain-specific foundation model to provide domain knowledge, the vision and language branch are prompted into the specific domain, Meanwhile, through our proposed cross-modal low-rank adaptation, the domain adaptation potentials of VLMs are mined in a cross-modal update manner.   }
	\label{main}
\end{figure*}

\section{Overview}
\subsection{Problem Setting}
The widely-used CLIP model is employed as the original natural vision-based VLM for the proposed generalized domain prompt learning framework. The CLIP model comprises a visual encoder and a text encoder, generating image embeddings and corresponding text embeddings, respectively. Built upon the vision transformer architecture, the CLIP model follows the same configuration as established in previous studies~\cite{khattak2023maple, zhou2022conditional}. During training, the CLIP model maximizes the cosine similarity between an image and its corresponding text and minimizes the cosine similarity between the image and unmatched text. This training strategy enables the CLIP model to perform zero-shot classification. For zero-shot classification, text embeddings $\omega_i$ are derived from predefined prompts, such as ``a photo of $category$", where $category$ represents the $i$-th class name. With $C$ categories and the visual embedding of the image denoted as $x$, the probability of the image belonging to the $i$-th class is calculated as:

\begin{equation}
\label{eq3}
p(y_i|x) = \frac{{\exp (sim(x,\omega_i )/\tau )}}{{\sum\nolimits_{i = 1}^C {\exp (sim(x,\omega_i )} /\tau )}},
\end{equation}
where $sim()$ denotes the cosine similarity function and $\tau$ represents the temperature hyperparameter. In prompt learning, the CLIP would be fine-tuned with some prompts so that the zero-shot classification ability on the downstream datasets gets boosted.  Previous prompt learning tasks mainly focus on natural-vison tasks, while our proposed generalized prompt learning provides domain-specific prompts, and transfers the powerful zero-shot classification ability of CLIP into specific domains.

\subsection{Generalized Domain Prompt Learning}
The generalized domain prompt learning aims to prompt VLMs efficiently from the constrained natural vision domain to the targeted specific domains, which provides accessible domain-specific VLMs for academia and advances VLM research for science domains.  To achieve this,  we leverage the domain-specific foundation model to provide the domain-specific knowledge. This knowledge successfully transfers the language and vision branches into the specific domain with the quaternion networks as the pipeline. Simultaneously, a fresh low-rank adaptation approach is proposed to fully exploit the domain adaptation potentials of VLMs. The overview of our proposed method is illustrated in Figure~\ref{main}. More specifically, the domain-specific images are first propagated into the domain-specific model and the following domain-project layer to obtain the domain-specific vision features. Subsequently, through simultaneously propagating the domain-specific vision features and the learnable context vector into the quaternion hidden space, the learnable context vectors are transferred from the constrained natural vision into the specific domain, which further generates the domain-specific text embeddings through concatenating with the category embeddings.  Moreover, based on the well-matched vision-language relationship,  the domain-specific vision features and the predefined domain-specific language prompts are jointly inputted into the quaternion networks to acquire the domain-specific vision prompts hierarchically. Consequently, domain-specific knowledge is propagated into the vision branch, transitioning it from the natural vision domain to a specific domain. Moreover, to fully exploit the domain-transfer potentials of VLMs, we set learnable low-rank weight matrixes to adjust the domain-specific features in each branch. To ensure a stable vision-language relationship, critical central low-rank weight matrices between the two branches are constrained to be simultaneously updated in a cross-modal manner. Consequently, our proposed generalized prompt learning successfully transfers the natural vision-based VLMs into specific domains and achieves remarkable performance in various domains.

\section{Results}\label{sec2}
\subsection{Datasets and Evaluation}
To evaluate our proposed method, we employ diverse datasets across multiple domains. In remote sensing, we utilize eight datasets: MLRSNet~\cite{qi2020mlrsnet} (109,161 images, 46 classes), PatternNet~\cite{zhou2018patternnet} (38 classes, 800 images/class), RSSCN7~\cite{zou2015deep} (7 classes, 400 images/class), AID~\cite{xia2017aid} (10,000 images, 30 classes), RSICD~\cite{lu2017exploring} (10,921 images, 30 classes), UCM~\cite{yang2010bag} (21 classes, 100 images/class), WHURS19~\cite{Dai2011WHURS19} (19 classes, 50 images/class), and NWPU~\cite{cheng2017remote} (40 classes, 700 images/class). For medical imaging, we use BTMRI~\cite{BTMRI} (7023 MRI images, 4 classes), CCBTM~\cite{CCBTM} (3264 MRI images, 4 classes), and CHMNIST~\cite{CHMNIST} (histology tiles, 8 classes). Geology analysis involves Berea sandstone~\cite{neumann202sandstones}, Doddington sandstone~\cite{moon2019intergranular}, Estaillade carbonate~\cite{muljadi2015estaillades}, Ketton carbonate~\cite{raeini2017generalized}, and Sandy multiscale medium ~\cite{mohammadmoradi2017multiscale}datasets. Fluid dynamics experiments use DNS data at $Re_{\lambda} \approx 250$, with 512×512 flow slice images collected every 2000 steps from 30000 to 50000 steps. In SAR, we employ MSTAR~\cite{MSTAR}, FUSAR-ship~\cite{hou2020fusar}, and SAR-ACD~\cite{SAR-ACD}. More details with regard to datasets can be found in the supplementary materials.

In the generalized domain prompt learning framework, the frozen domain-specific foundation models are leveraged to provide the domain knowledge. For the remote sensing domain, we adopt the remote sensing foundation models~\cite{wang2022advancing} including ViT~\cite{dosovitskiy2020image} and ViTAE~\cite{xu2021vitae} to conduct experiments. For the medical domain, we leverage the encoder of the popular MedicalSAM~\cite{ma2023segment} as the medical foundation model to provide medical domain knowledge. However, there are no available open-source domain-specific foundation models for the geology, SAR, and fluid dynamics domains. Therefore, we obtain domain-specific foundation models for these domains by pre-training the plain ViT using the domain datasets in a masked autoencoder (MAE)~\cite{he2022masked} manner. Specifically, in our experiments, the Sandy multiscale medium~\cite{mohammadmoradi2017multiscale} is adopted as the pre-trained dataset to adopt the geology foundation models. Meanwhile, we use the open-source SAR-OOD dataset~\cite{SAR-OOD} as the pre-trained dataset to acquire the SAR foundation model.  As for the fluid dynamics domain, we respectively utilize the remaining flow images at other steps as the pre-trained datasets to acquire three distinct fluid foundation models for 30000-step, 40000-step, and 50000-step.

As a prompt learning task for the generalized domain, we utilized the same evaluation metrics in previous natural vision-based methods~\cite{radford2021learning,zhou2022learning,zhou2022conditional,khattak2023maple}. Namely, the generalized classification task: with the given category names of the whole datasets and a few training samples of base categories(16 shots following the previous setting) as the prompts, the advanced models should successfully classify each input into their correct category of both seen base categories and unseen novel categories.   The classification accuracy ($Acc_{base}$ for base classes and $Acc_{novel}$ for novel categories)  is utilized as the main evaluation metric. Like the previous method. Harmonic Mean (HM) score is also leveraged as evaluation metrics:
\begin{equation}
    HM = \frac{{2 \times Acc_{base} \times Acc_{novel}}}{{Acc_{base} + Acc_{novel}}},
\end{equation}
where $Acc_{base}$ denotes the accuracy of base categories, and $Acc_{novel}$ denotes the accuracy of novel categories. In experiments, We compare our method with previous SOTA nature vision-based prompt learning methods to ensure a fair comparison. Our superior performance could demonstrate that the obvious domain gap seriously demotes the performance, and our method effectively achieves the generalized domain transfer of VLMs.

\begin{figure*}[!t]
	\begin{center}
		\includegraphics[width=1.0\linewidth]{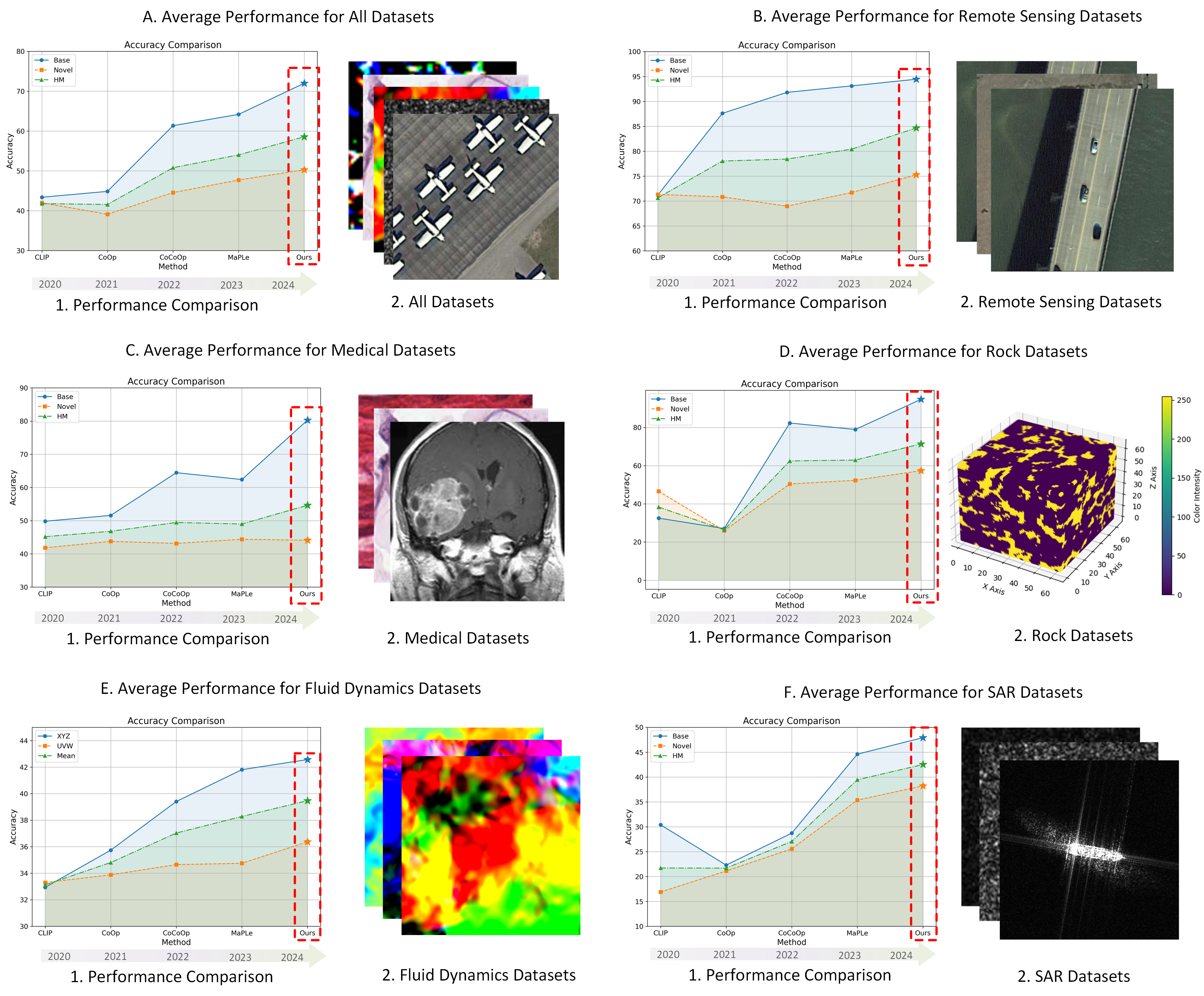}
	\end{center}
	\caption{Comparison between our method and SOTA methods in terms of average performance. Our method performs well over the compared methods for five domains. We use \red{red} frame to highlight our performance. HM denotes the harmonic mean score, and the others are the accuracy scores.  }
	\label{fig_main}
\end{figure*}

\subsection{Evaluation Results}

Through the generalized domain prompt learning, we provide domain-specific VLMs across various domains, and the average evaluation performances are summarized in Figure~\ref{fig_main}. Obviously, our generated domain-specific VLMs achieve advanced performance across different domains. The detailed experimental results are illustrated in the supplementary materials for reference.  

\noindent \textbf{Remote Sensing VLM.} As presented in Figure~\ref{fig_main} (B), the acquired remote sensing VLM surpasses previous nature vision-based prompt learning methods. This enhancement results in an approximate 3\% mean HM performance improvement. This underscores the effectiveness of utilizing domain-specific foundation models to guide the transfer of natural vision-based VLMs into remote sensing. Notably, despite ViTAE's greater depth compared to ViT, our ViT-based method achieves superior overall performance. This observation suggests that excessively deep architectures may not be optimal for extracting domain-specific information, as surplus parameters could potentially hinder performance.

\noindent \textbf{Medical VLM.} The performance improvements within the medical domain, as detailed in  Figure~\ref{fig_main} (C), are noteworthy. Our method consistently outperforms previous state-of-the-art nature vision-based prompt learning techniques. Particularly striking is the remarkable 5.19\% boost in mean Harmonic Mean (HM), coupled with a substantial rise in base accuracy from 64.45\% to 80.27\%. Despite these notable successes, our investigation uncovers an intriguing observation: our method doesn't achieve optimal performance in novel categories. This suggests a limitation in the medical foundation model's ability to effectively generalize to unseen categories during prompt learning.

\noindent \textbf{Geology VLM.} We validate our proposed domain prompt learning framework within the field of geology, using rock imagery as examples, with results summarized in  Figure~\ref{fig_main} (D). Our method consistently outperforms previous prompt learning algorithms by a significant margin across various performance metrics: 12.41\% for base categories, 5.05\% for novel classes, and 8.36\% for the Harmonic Mean (HM). These findings underscore the critical role of introduced domain knowledge in enhancing rock image recognition, affirming the effectiveness of our method in seamlessly transferring natural vision-based Vision-Language Models (VLMs) into the geology domain.

\noindent \textbf{Fluid Dynamics VLM.} 
Using fluid simulation based on the Navier-Stokes equation, we generated a series of fluid datasets, and the experimental results are presented in Figure~\ref{fig_main} (E). The task of accurately capturing minute-scale structures within the flow field poses a significant challenge for differentiation. Statistical significance across various directions is minimal, with differentiation achievable solely through discerning fluidity in intricate structures. Despite the modest recognition performance observed across all prompt learning methods, our proposed generalized domain prompt learning approach, strengthened by domain-specific foundation models, significantly enhances the recognition of flow images. Notably, our method demonstrates improvements in average performance, with a 0.76\% refinement in classifying XYZ dimensions, a 1.62\% improvement in UVW dimension classification, and a 1.19\% boost in mean performance. These findings underscore the efficacy of our approach in advancing flow image recognition, highlighting the pivotal role of domain-specific knowledge in prompt learning tasks.

\noindent \textbf{SAR VLM.} The average evaluation performance of SAR VLM is presented in Figure~\ref{fig_main} (F). Interestingly, we observe that both CoOp and CoCoOp consistently achieve similar or even inferior performance compared to CLIP. In contrast, our method consistently outperforms the compared methods, showcasing robust performance across base and novel categories. These observations collectively demonstrate the efficacy of our approach in seamlessly transferring vision-based VLMs into the SAR domain, providing an effective SAR VLM.

\begin{figure*}[!t]
	\begin{center}
		\includegraphics[width=1.0\linewidth]{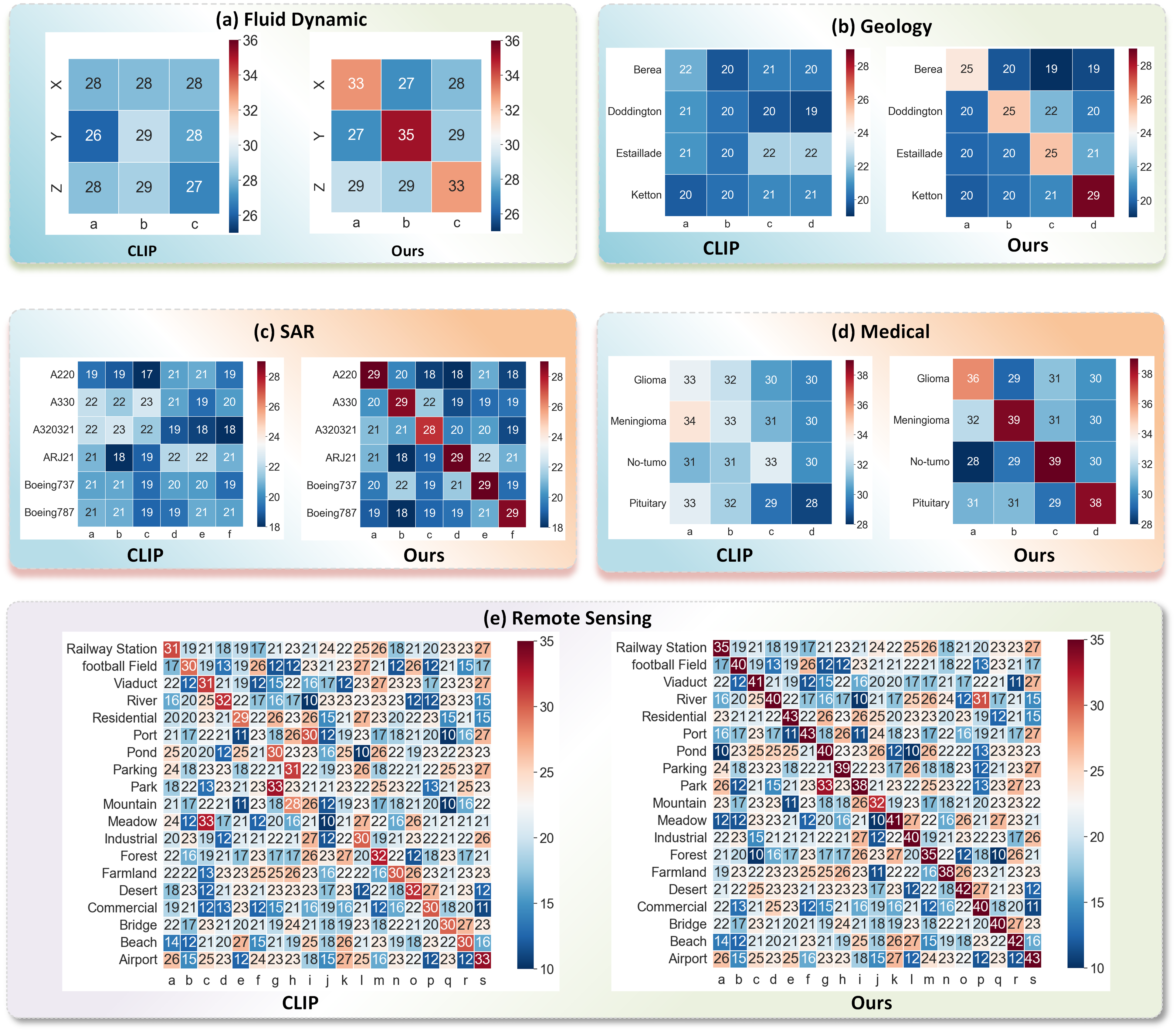}
	\end{center}
	\caption{Visualization for different datasets. The scores denote the similarity scores.  Higher similarity scores indicate that the sample is more likely to belong to the corresponding category.  }
	\label{fig_vis}
\end{figure*}

\subsection{Case Study.} 
\textbf{Qualitative Comparison.} The training process of our method employs contrastive learning loss to supervise prompt learning, while the inference process adopts a zero-shot classification approach. In this manner, given only the names of all categories, input images are classified into the most similar class by computing cosine similarities between language features and image features. The objective of prompt learning is to maximize the similarity between input images and corresponding category name-based language features while minimizing other similarities. To directly visualize the performance differences, we collect computed cosine similarities for different categories across various datasets: 3000-step dataset for the fluid dynamics domain, Rock X dataset for the geology domain, SAR-ACD dataset for the SAR domain, BTMRI dataset for the medical domain, and WHURS19 dataset for the remote sensing domain. The overall comparisons are presented in Figure~\ref{fig_vis}. It is evident that while the original CLIP often misclassifies input images due to close or higher similarities for incorrect categories, our method rectifies these errors by suppressing similarities for wrong categories and amplifying similarities for true classes. Furthermore, a deeper analysis reveals that as the number of categories increases, the gap between similarities for true and false classes widens for both CLIP and our method. This phenomenon suggests that in contrastive learning, creating more separate spaces can lead to improved recognition performance.

\begin{figure*}[!t]
	\begin{center}
		\includegraphics[width=1.0\linewidth]{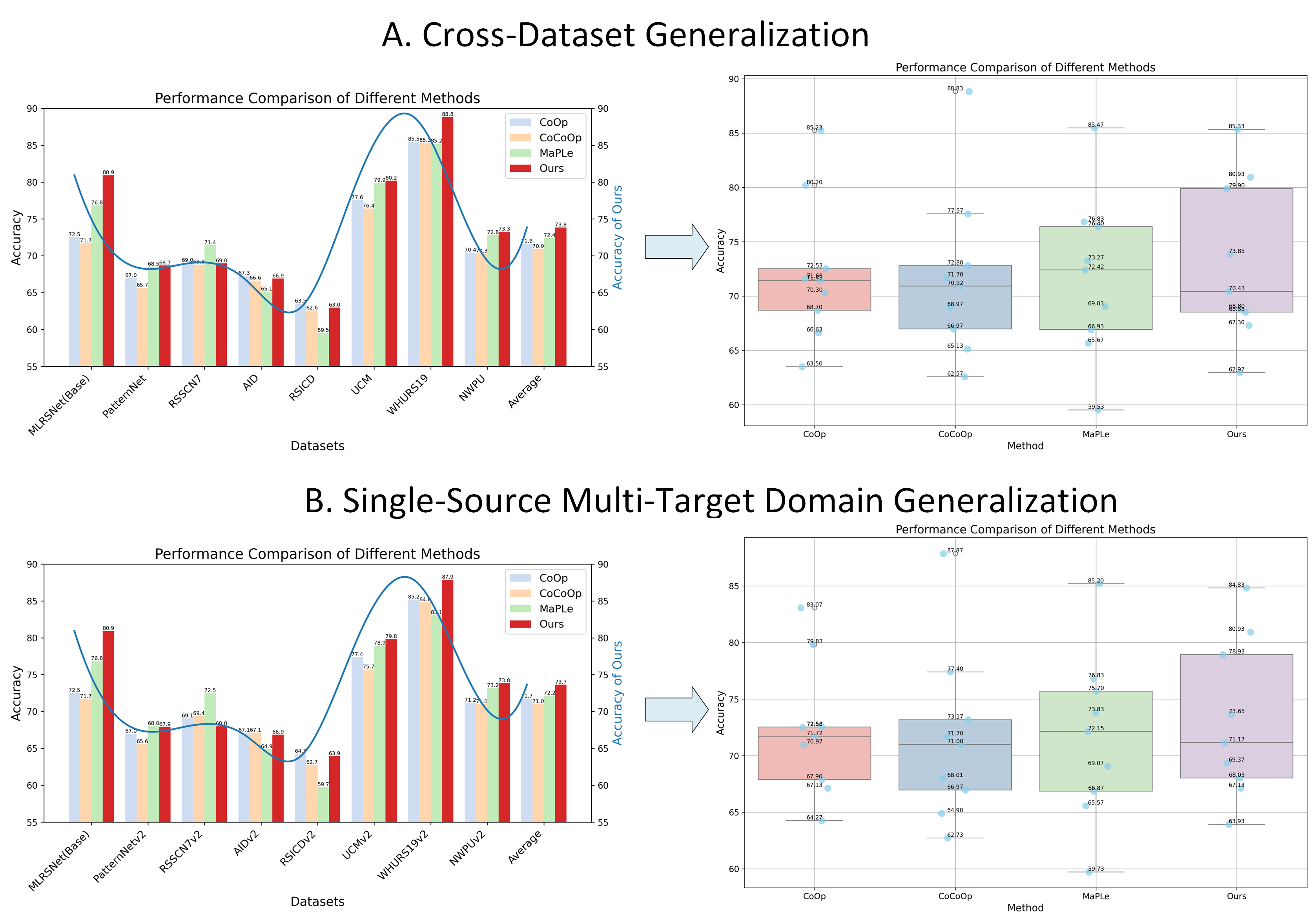}
	\end{center}
	\caption{Comparisons with SOTA methods for (A) cross-dataset generalization and (B) single-source multi-target domain generalization with the MLRSNet dataset as the source dataset. The proposed method achieves better performance than the compared methods.  }
	\label{fig_gen}
\end{figure*}

\noindent \textbf{Cross-Dataset Evaluation.} To assess the cross-dataset capability of our method, we adopt the evaluation setting established by previous studies~\cite{khattak2023maple}.  Specifically, we conduct cross-dataset evaluations using remote sensing datasets as examples, with experimental results summarized in Figure~\ref{fig_gen} (A). Detailed experimental results can be found in supplementary materials.  Training is performed on the MLRSNet dataset, and model performance is evaluated on the base categories of all datasets. Although our method may not achieve the highest performance on certain datasets, the mean performance consistently surpasses that of other advanced methods, exhibiting a noteworthy 1.43\% improvement. Notably, our method achieves its highest performance on the WHURS19 dataset, with an accuracy of 88.83\% and a remarkable 3.36\% performance gain. These findings suggest that our proposed prompt learning method demonstrates relatively superior cross-dataset recognition performance, showcasing its adaptability across diverse datasets.

\noindent \textbf{Category Generalization.} Following the methodology established by previous prompt learning studies~\cite{khattak2023maple} within the category generalization setting, we further explore the generalization capabilities of our proposed method. Specifically, we evaluate the MLRSNet-based model on all categories of all datasets, with the experimental results detailed in Figure~\ref{fig_gen} (B). To distinguish this evaluation from the previous one, we append 'V2' to the dataset names. Once again, our method outperforms previous approaches in terms of average performance, demonstrating a notable 1.50\% improvement. Additionally, our method achieves the highest performance on the WHURS19v2 dataset. However, we observe a failure in performance on the RSSCN7v2 dataset. Further examination of this phenomenon reveals that the limited number of categories in the RSSCN7v2 dataset may account for this discrepancy. This finding aligns with our observation in the Qualitative Comparison section, suggesting that a restricted learning space could indeed hinder recognition performance.

\begin{figure}[h!]
  \centering
  \begin{subfigure}[b]{0.45\linewidth}
    \includegraphics[width=\linewidth]{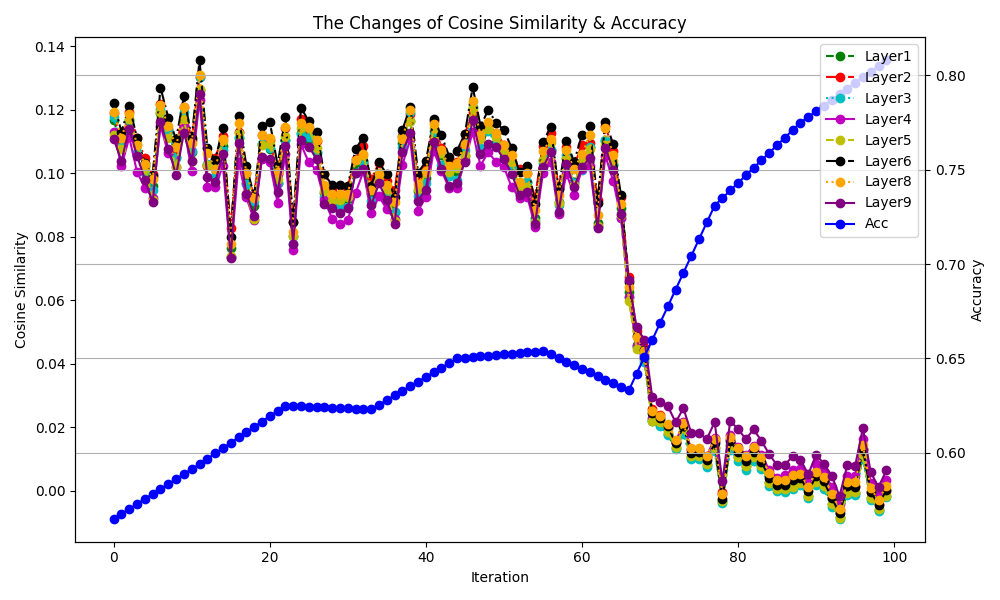}
    \caption{Orthogonality Study on RSICD dataset.}
    \label{fig:sub1}
  \end{subfigure}
  \begin{subfigure}[b]{0.45\linewidth}
    \includegraphics[width=\linewidth]{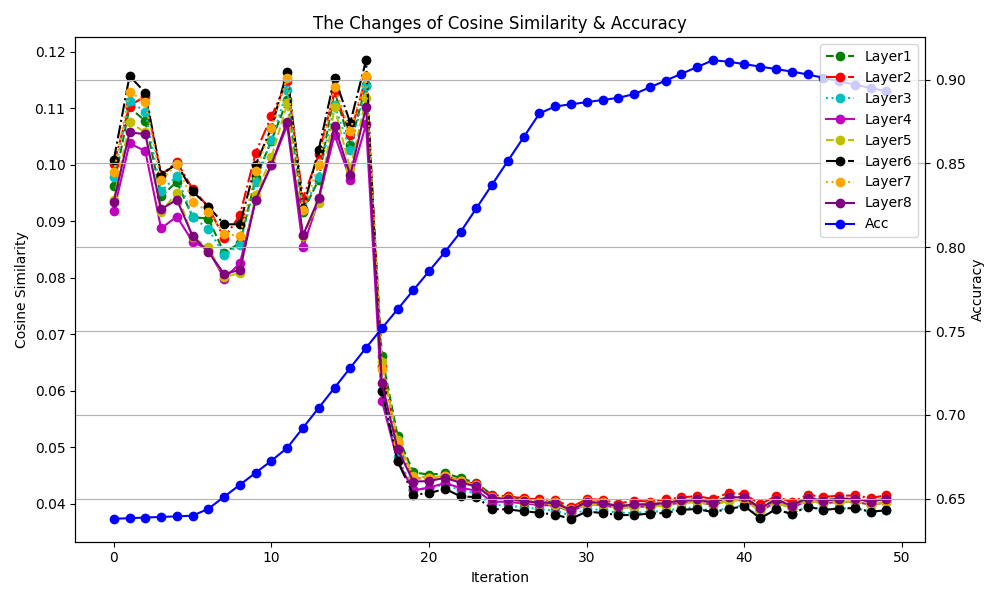}
    \caption{Orthogonality Study on RSCCN7 dataset.}
    \label{fig:sub2}
  \end{subfigure}
  \caption{The Study of Orthogonality.}
  \label{fig:combined}
\end{figure}

\noindent \textbf{The Study of Orthogonality.} In our proposed framework for generalized domain prompt learning, we harness quaternion networks to uncover the latent orthogonal relationship between multi-modal features. To investigate the significance of orthogonality, we conducted experiments on the RSICD and RSSCN7 datasets as case studies. We tracked the cosine similarities among multi-modal inputs at different network layers and monitored changes in accuracy, as depicted in Figure~\ref{fig:combined}. Remarkably, we observed a consistent decrease in cosine similarities approaching zero with iterative refinement. Specifically, for the RSICD dataset, this decline occurred gradually, with the most significant drop happening around the 65th iteration. Conversely, for the RSSCN7 dataset, a rapid decline was observed much earlier, around the 15th iteration. Notably, as cosine similarities decrease, test accuracy exhibits corresponding improvement. Particularly intriguing is the rapid increase in accuracy coinciding with sharp declines in cosine similarity. This phenomenon underscores the strong correlation between effectively identified orthogonality and superior performance. Our findings demonstrate that leveraging quaternion networks within the generalized domain prompt learning framework facilitates the extraction of orthogonal relationships between multi-modal inputs, thereby enhancing domain-recognition performance.

\section{Discussion}

In our study, we propose generalized domain prompt learning methods that successfully prompt the strong zero-shot recognition ability of VLMs from the natural vision domain to the target-specific domains. The generalized domain prompt learning provides accessible VLMs for diverse science domains with only a few annotated samples as prompts, advancing sustainable and equitable VLM research for academia. Particularly, our proposed method leverages domain-specific foundation models for integrating domain-specific knowledge into both visual and language modalities. Additionally, we introduce a pioneering cross-modal low-rank adaptation method to harness the domain adaptation potential of VLMs through constrained cross-modal updates. To assess the efficacy of our approach, we conduct experiments across diverse domains including remote sensing, medical imaging, geology, SAR, and fluid dynamics. Our results consistently demonstrate superior performance across all domains, suggesting that our method effectively bridges the recognition gap between the constrained natural vision domains and specific domains, thereby opening new avenues for leveraging powerful VLMs.

Interestingly, while our cross-modal low-rank adaptation significantly boosts the domain adaptation capabilities of VLMs, we observe that further performance improvements stem from incorporating domain knowledge from domain-specific foundation models. This underscores the paramount importance of domain knowledge in the domain prompt learning process. Future research directions could explore more effective methods of integrating domain knowledge for enhanced domain prompt learning. Moreover, our findings indicate that while our approach enhances performance on unseen novel categories, it yields even greater improvements for base categories. This suggests that specific examples within categories can significantly aid the model in understanding domain-specific nuances. Furthermore, our investigation into the orthogonality aspect reveals a noteworthy trend during the training process: the cosine similarities between multi-modal inputs gradually approach zero, while test accuracy experiences a rapid increase in tandem with this decline. This observed phenomenon underscores the pivotal role of orthogonality in effectively modeling the relationships between multi-modal inputs, consequently enhancing domain recognition performance. Thus, we posit that refining the modeling of orthogonality between multi-modal inputs holds significant promise for advancing the quality of domain-aware multi-modal features.

Our study primarily focuses on addressing domain-specific recognition challenges within the framework of generalized domain prompt learning, achieving impressive performance. Given the foundational importance of recognition, the success of prompted domain-specific VLMs in domain-specific recognition suggests promising avenues for extending domain prompt learning to other critical tasks for science domains such as language-driven science object generation and object detection in future research endeavors.

\section{Methods}
\subsection{Preliminaries}

\textbf{Quaternion Modeling.} In four-dimensional space, a quaternion $Q$ extends beyond a complex number and can be denoted as:
\begin{equation}
 Q= r1+ x\mathbf{i} + y\mathbf{j} + z\mathbf{k},
\end{equation}
where $r$, $x$, $y$, and $z$ are real numbers, and $1$, $\mathbf{i}$, $\mathbf{j}$, and $\mathbf{k}$ are the quaternion unit basis. Here, $r$ represents the real part of $Q$, while $x\mathbf{i} + y\mathbf{j} + z\mathbf{k}$ denotes its imaginary or vector part. By projecting the multi-modal features into the quaternion field, the quaternion network could efficiently mine the orthogonal relations between multi-modal features. More details about the quaternion networks could be found in the supplementary materials. 



\noindent\textbf{Low-Rank Adaptation.} Low-Rank Adaptation (LoRA)~\cite{hu2021lora} introduced a novel mechanism involving a learnable low-rank matrix $\omega_L$, which operates concurrently with the original weight matrix $\omega \in \mathbb{R}^{D \times K}$. This original weight matrix is typically associated with the parameter matrix in the multi-head self-attention module of each transformer layer, as proposed in~\cite{hu2021lora}. The learnable low-rank matrix $\omega$ is derived through a QR decomposition, represented as $\omega_L = \beta \alpha$, where $\beta \in \mathbb{R}^{D \times V}$ and $\alpha \in \mathbb{R}^{V \times K}$. This innovative approach introduces a new dimension to the adaptation process, enhancing the model's flexibility and adaptability.

\noindent\textbf{Prompt Learning.} The CLIP model comprises a visual encoder and a text encoder, generating image embeddings and corresponding text embeddings, respectively. Based on the vision transformer, the CLIP model operates under the same framework as previous methods~\cite{khattak2023maple, zhou2022conditional}. During training, it maximizes the cosine similarity between the image and its matched text while minimizing the similarity with unmatched text, facilitating zero-shot classification.
For zero-shot classification, the text embedding $\omega_i$ is derived from a predefined prompt, such as "a photo of $category$", where $category$ denotes the $i$-th class name. If there are $C$ categories and the visual embedding of the image is $x$, the probability of the image belonging to the $i$-th class name is computed as:
\begin{equation}
\label{eq3}
p(y_i|x) = \frac{{\exp (sim(x,\omega_i )/\tau )}}{{\sum\nolimits_{i = 1}^C {\exp (sim(x,\omega_i )} /\tau )}},
\end{equation}
where $sim()$ represents the cosine similarity function and $\tau$ is the temperature hyperparameter.

\subsection{Prompting Language Branch}
Firstly, to obtain the domain-specific vision feature $F_d$, the image patch embeddings are propagated into the corresponding domain-specific foundation model.  Subsequently, 
to prepare the appropriate domain-specific information for the computation in the quaternion hidden space,  a domain-projection layer $L_d$ consisting of two linear neural layers is applied to produce the project domain-specific vision features $\widehat F_d$:
\begin{equation}
{\widehat F_d} = {L_d}({F_d}),
\end{equation}

In order to mine the critical orthogonal intermodal relationship, the domain-specific vision features $\widehat F_d$ and the learnable context embeddings $E_c$ are modelled in two orthogonal axes in the quaternion hidden space:
\begin{equation}
Q_l = E_c + {\widehat F_d}\textbf{i} + 0\textbf{j} + 0\textbf{k},
\end{equation}

To facilitate quaternion projection in the quaternion hidden space, we create a zero tensor $Z_0$ with the same dimensions as $E_c$. Additionally, in line with the observations in CoCoOp~\cite{zhou2022conditional}, prompt learning often encounters overfitting issues due to the limited data available. To mitigate this concern, we introduce random Gaussian noise $N_G$ into the quaternion hidden space, aiming to implicitly bolster the robustness of the learning process. The magnitude of this noise is scaled by the mean of the domain-specific vision features $\widehat F_d$, defined as:
\begin{equation}
    N_G = \text{Mean}(\widehat F_d) \cdot N_{\theta},
\end{equation}
where $N_{\theta}$ represents standard Gaussian noise. Formally, for a given quaternion layer $Q_t$, the computation of the domain-specific context embedding $T_d$ is expressed as:
\begin{equation}
    T_d = Q_t([\widehat F_d + E_c + N_G, Z_0]),
\end{equation}
By employing this approach, we effectively extract orthogonal intermodal relationships between domain-specific vision features and contextual embeddings within the quaternion hidden space, enabling the projection of generalized contextual embeddings into the specific domain space.

To incorporate domain-specific information into the language encoder in a prompt manner, we utilize the domain-specific context embedding $T_d$ and introduce a set of learnable language prompt features $\left[P_l^1, P_l^2,..., P_l^m\right]$ with a depth setting of $k$ into the encoder layers $\left[L_l^1, L_l^2,..., L_l^m\right]$ of the language branch, where $m$ represents the total number of layers. Given the fixed text embedding $C_t$ for categories, we first concatenate $T_d$ with $C_t$ to obtain the complete domain-specific language embeddings, denoted as $E^l_1 = \left[T_d, C_t \right]$. The propagation of domain-specific information is then computed as follows:
\begin{equation}
\left[ {E^l_i,~~\_~~} \right] = L_l^i([E^l_{i-1},P_l^{i-1}])~~~~~\text{for}~~i = 1,...,k,
\end{equation}
\begin{equation}
\left[ {E^l_j,P_l^j} \right] = L_l^j([E^l_{j-1},P_l^{j-1}])~~~~~\text{for}~~j = k+1,...,m,
\end{equation}

\subsection{Prompting Vision Branch}

The pre-trained VLMs establish a robust vision-language matching relationship, facilitating the seamless transfer of domain-specific information from the language branch to the vision branch. To enable this transfer, we introduce a set of learnable vision prompt features, represented as $\left[P_v^1, P_v^2,..., P_v^m\right]$, with a depth of $k$, corresponding to the encoder layers $\left[L_v^1, L_v^2,..., L_v^m\right]$ of the vision branch. Additionally, we introduce a series of quaternion layers $\left[Q_v^1, Q_v^2,..., Q_v^m\right]$, also with a depth of $k$, tasked with providing cross-modal domain-specific information.

Similar to the quaternion computation in the language branch, domain-specific vision features $\widehat F_d$ and language prompt features $P_l^i$ are modeled along two orthogonal axes in the quaternion hidden space for the vision branch. This computation is expressed as follows:
\begin{equation}
Q_v^i = P_l^i + {\widehat F_d}\mathbf{i} + 0\mathbf{j} + 0\mathbf{k},
\end{equation}

Given the well-matched vision-language relationship, random noise is deemed unnecessary. Thus, the computation of vision prompt features is simplified to:
\begin{equation}
    P_v^i = Q_v^i([\widehat F_d + P_l^i, Z_0])~~~~~~\text{for}~~i = 1,2,...,k,
\end{equation}

By leveraging the well-established vision-language relations, we effectively propagate domain-specific knowledge into the vision prompt features within the quaternion hidden space. To propagate domain-specific information within the vision branch, we commence with the original image embeddings $E^v_1$ and class token $c_1$, computing the propagation as follows:
\begin{equation}
\left[ {E^v_i, c_i, \_} \right] = L_v^i([E^v_{i-1}, c_{i-1}, P_v^{i-1}]) ~~~\text{for}~~i = 1,...,k,
\end{equation}

Subsequently, we further propagate the domain-specific information into the Cascading language branches using the following equation:
\begin{equation}
\left[ {E^v_j, c_i, P_v^j} \right] = L_v^j([E^v_{j-1}, c_{j-1}, P_v^{j-1}]) ~~~\text{for}~~ j = k+1,...,m,
\end{equation}

\subsection{Cross-Modal Low-Rank Adaptation}

To fully exploit the domain-adaptation potential of Vision-Language Models (VLMs), we aim to extend the well-established vision-language relationship from a generalized to a specialized context. While the original Low-Rank Adaptation (LoRA) method effectively adapts pretrained models to target downstream tasks with limited samples, it overlooks the crucial vision-language relationship. Hence, we propose a novel approach, termed cross-modal Low-Rank adaptation, to ensure stable domain adaptation of VLMs while preserving the vision-language relationship.

For each layer of the language branch, we introduce learnable weights $\omega^{l,i}_L \in \mathbb{R}^{D \times K}$. Unlike the original LoRA weights $\omega_L$, our LoRA weights comprise three learnable components: $\omega^{l,i}_L = \beta^{l,i} \Lambda^{l,i} \alpha^{l,i}$, where $\beta^{l,i} \in \mathbb{R}^{D \times V}$, $\alpha^{l,i} \in \mathbb{R}^{V \times K}$, and $\Lambda^{l,i}$ is a learnable diagonal matrix defined as:
\begin{equation}
\Lambda^{l,i} = \begin{bmatrix} 
\lambda^{l,i}_1 & \cdots & 0 \\ 
\vdots & \ddots & \vdots \\ 
0 & \cdots & \lambda^{l,i}_V 
\end{bmatrix} \in \mathbb{R}^{V \times V}.
\end{equation}

Similarly, we introduce learnable weights $\omega^{v,i}_L \in \mathbb{R}^{D \times K}$ for each layer of the vision branch. To maintain a stable vision-language relationship, we constrain $\Lambda^{l,i}$ and $\Lambda^{v,i}$ to be updated through matrix multiplication. Specifically, we introduce a learnable cross-modal matrix ${M_c} \in {\mathbb{R}^{2 \times 2}}$. Then, for each layer of the vision and language branch, $\Lambda^{l,i}$ and $\Lambda^{v,i}$ are updated in dimension using $M_c$ to obtain updated cross-modal matrices $\widehat \Lambda^{l,i}$ and $\widehat \Lambda^{v,i}$:
\begin{equation}
[\widehat \Lambda ^{l,i}, \widehat \Lambda ^{v,i}] = {M_c} \otimes [\Lambda ^{l,i}, \Lambda ^{v,i}],
\end{equation}
where $[\cdot,\cdot]$ denotes the concatenation operation, and $\otimes$ represents matrix multiplication. Correspondingly, the cross-modal LoRA weights are updated as follows:
\begin{equation}
\widehat \omega^{l,i}_L = \beta^{l,i} \widehat \Lambda^{l,i} \alpha^{l,i}, ~~~~\widehat \omega^{v,i}_L = \beta^{v,i} \widehat \Lambda^{v,i} \alpha^{v,i}.
\end{equation}

Subsequently, with the cross-modal LoRA weights as the weights of linear layers, domain-shift features can be computed and directly added to the image and text embeddings as follows:
\begin{equation}
\widehat E^l_i = E^l_i + \widehat \omega^{l,i}_L(E^l_{i-1}) ~~~\text{for}~~i = 2,...,m,
\end{equation}
\begin{equation}
\widehat E^v_i = E^v_i + \widehat \omega^{v,i}_L(E^v_{i-1}) ~~~\text{for}~~i = 2,...,m.
\end{equation}

This approach effectively leverages the domain adaptation potential of VLMs and ensures the transfer of the well-matched vision-language relations from the generalized to the specific domain space.

\subsection{Final Classification}

Finally, given the vision embeddings $E^v_m$ from the last layer of the vision branch and the language embeddings $\left[E^{l,1}_m,E^{l,2}_m,..., E^{l,C}_m\right]$ for $C$ categories from the last layer of the language branch, we compute the probability of an image belonging to a specific category as follows:  
\begin{equation}
p(y_i|x) = \frac{{\exp \left(\text{sim}\left(E^v_m,E^{l,i}_m\right)/\tau \right)}}{{\sum\nolimits_{j = 1}^C {\exp \left(\text{sim}\left(E^v_m,E^{l,j}_m\right)/\tau \right)} }},
\end{equation}
where $p(y_i|x)$ represents the probability of the image belonging to the $i$-th category, $\text{sim}$ denotes the similarity function, and $\tau$ is a temperature parameter.

\section{Data availability}
Datasets for the remote sensing domain are accessible through their respective online sources. All datasets utilized in the medical domain can be obtained from https://www.kaggle.com/. The rock images utilized in this study were gathered from previously mentioned online references. Flow images for the fluid dynamics domain were generated through the simulation of the Navier-Stokes equation and will be made available at https://github.com/caoql98/GDPL. Additionally, all SAR datasets utilized in this research were collected from their corresponding online sources.

\section{Code Availability }
The code used for training, testing the model, and generating new datasets will be accessible at https://github.com/caoql98/GDPL.









\bibliography{sn-bibliography}


\begin{thebibliography}{47}
\ifx \bisbn   \undefined \def \bisbn  #1{ISBN #1}\fi
\ifx \binits  \undefined \def \binits#1{#1}\fi
\ifx \bauthor  \undefined \def \bauthor#1{#1}\fi
\ifx \batitle  \undefined \def \batitle#1{#1}\fi
\ifx \bjtitle  \undefined \def \bjtitle#1{#1}\fi
\ifx \bvolume  \undefined \def \bvolume#1{\textbf{#1}}\fi
\ifx \byear  \undefined \def \byear#1{#1}\fi
\ifx \bissue  \undefined \def \bissue#1{#1}\fi
\ifx \bfpage  \undefined \def \bfpage#1{#1}\fi
\ifx \blpage  \undefined \def \blpage #1{#1}\fi
\ifx \burl  \undefined \def \burl#1{\textsf{#1}}\fi
\ifx \doiurl  \undefined \def \doiurl#1{\url{https://doi.org/#1}}\fi
\ifx \betal  \undefined \def \betal{\textit{et al.}}\fi
\ifx \binstitute  \undefined \def \binstitute#1{#1}\fi
\ifx \binstitutionaled  \undefined \def \binstitutionaled#1{#1}\fi
\ifx \bctitle  \undefined \def \bctitle#1{#1}\fi
\ifx \beditor  \undefined \def \beditor#1{#1}\fi
\ifx \bpublisher  \undefined \def \bpublisher#1{#1}\fi
\ifx \bbtitle  \undefined \def \bbtitle#1{#1}\fi
\ifx \bedition  \undefined \def \bedition#1{#1}\fi
\ifx \bseriesno  \undefined \def \bseriesno#1{#1}\fi
\ifx \blocation  \undefined \def \blocation#1{#1}\fi
\ifx \bsertitle  \undefined \def \bsertitle#1{#1}\fi
\ifx \bsnm \undefined \def \bsnm#1{#1}\fi
\ifx \bsuffix \undefined \def \bsuffix#1{#1}\fi
\ifx \bparticle \undefined \def \bparticle#1{#1}\fi
\ifx \barticle \undefined \def \barticle#1{#1}\fi
\bibcommenthead
\ifx \bconfdate \undefined \def \bconfdate #1{#1}\fi
\ifx \botherref \undefined \def \botherref #1{#1}\fi
\ifx \url \undefined \def \url#1{\textsf{#1}}\fi
\ifx \bchapter \undefined \def \bchapter#1{#1}\fi
\ifx \bbook \undefined \def \bbook#1{#1}\fi
\ifx \bcomment \undefined \def \bcomment#1{#1}\fi
\ifx \oauthor \undefined \def \oauthor#1{#1}\fi
\ifx \citeauthoryear \undefined \def \citeauthoryear#1{#1}\fi
\ifx \endbibitem  \undefined \def \endbibitem {}\fi
\ifx \bconflocation  \undefined \def \bconflocation#1{#1}\fi
\ifx \arxivurl  \undefined \def \arxivurl#1{\textsf{#1}}\fi
\csname PreBibitemsHook\endcsname

\bibitem[\protect\citeauthoryear{Simonyan and Zisserman}{2014}]{simonyan2014very}
\begin{botherref}
\oauthor{\bsnm{Simonyan}, \binits{K.}},
\oauthor{\bsnm{Zisserman}, \binits{A.}}:
Very deep convolutional networks for large-scale image recognition.
arXiv:1409.1556
(2014)
\end{botherref}
\endbibitem

\bibitem[\protect\citeauthoryear{He et~al.}{2016}]{he2016deep}
\begin{bchapter}
\bauthor{\bsnm{He}, \binits{K.}},
\bauthor{\bsnm{Zhang}, \binits{X.}},
\bauthor{\bsnm{Ren}, \binits{S.}},
\bauthor{\bsnm{Sun}, \binits{J.}}:
\bctitle{Deep residual learning for image recognition}.
In: \bbtitle{Proceedings of the IEEE Conference on Computer Vision and Pattern Recognition}
(\byear{2016})
\end{bchapter}
\endbibitem

\bibitem[\protect\citeauthoryear{Long et~al.}{2015}]{long2015fully}
\begin{bchapter}
\bauthor{\bsnm{Long}, \binits{J.}},
\bauthor{\bsnm{Shelhamer}, \binits{E.}},
\bauthor{\bsnm{Darrell}, \binits{T.}}:
\bctitle{Fully convolutional networks for semantic segmentation}.
In: \bbtitle{Proceedings of the IEEE Conference on Computer Vision and Pattern Recognition}
(\byear{2015})
\end{bchapter}
\endbibitem

\bibitem[\protect\citeauthoryear{Chen et~al.}{2017}]{chen2017deeplab}
\begin{barticle}
\bauthor{\bsnm{Chen}, \binits{L.-C.}},
\bauthor{\bsnm{Papandreou}, \binits{G.}},
\bauthor{\bsnm{Kokkinos}, \binits{I.}},
\bauthor{\bsnm{Murphy}, \binits{K.}},
\bauthor{\bsnm{Yuille}, \binits{A.L.}}:
\batitle{Deeplab: Semantic image segmentation with deep convolutional nets, atrous convolution, and fully connected crfs}.
\bjtitle{IEEE transactions on pattern analysis and machine intelligence}
\bvolume{40}(\bissue{4}),
\bfpage{834}--\blpage{848}
(\byear{2017})
\end{barticle}
\endbibitem

\bibitem[\protect\citeauthoryear{Redmon et~al.}{2016}]{redmon2016you}
\begin{bchapter}
\bauthor{\bsnm{Redmon}, \binits{J.}},
\bauthor{\bsnm{Divvala}, \binits{S.}},
\bauthor{\bsnm{Girshick}, \binits{R.}},
\bauthor{\bsnm{Farhadi}, \binits{A.}}:
\bctitle{You only look once: Unified, real-time object detection}.
In: \bbtitle{Proceedings of the IEEE Conference on Computer Vision and Pattern Recognition}
(\byear{2016})
\end{bchapter}
\endbibitem

\bibitem[\protect\citeauthoryear{Radford et~al.}{2021}]{radford2021learning}
\begin{bchapter}
\bauthor{\bsnm{Radford}, \binits{A.}},
\bauthor{\bsnm{Kim}, \binits{J.W.}},
\bauthor{\bsnm{Hallacy}, \binits{C.}},
\bauthor{\bsnm{Ramesh}, \binits{A.}},
\bauthor{\bsnm{Goh}, \binits{G.}},
\bauthor{\bsnm{Agarwal}, \binits{S.}},
\bauthor{\bsnm{Sastry}, \binits{G.}},
\bauthor{\bsnm{Askell}, \binits{A.}},
\bauthor{\bsnm{Mishkin}, \binits{P.}},
\bauthor{\bsnm{Clark}, \binits{J.}}, \betal:
\bctitle{Learning transferable visual models from natural language supervision}.
In: \bbtitle{International Conference on Machine Learning}
(\byear{2021})
\end{bchapter}
\endbibitem

\bibitem[\protect\citeauthoryear{Su et~al.}{2019}]{su2019vl}
\begin{botherref}
\oauthor{\bsnm{Su}, \binits{W.}},
\oauthor{\bsnm{Zhu}, \binits{X.}},
\oauthor{\bsnm{Cao}, \binits{Y.}},
\oauthor{\bsnm{Li}, \binits{B.}},
\oauthor{\bsnm{Lu}, \binits{L.}},
\oauthor{\bsnm{Wei}, \binits{F.}},
\oauthor{\bsnm{Dai}, \binits{J.}}:
Vl-bert: Pre-training of generic visual-linguistic representations.
arXiv:1908.08530
(2019)
\end{botherref}
\endbibitem

\bibitem[\protect\citeauthoryear{Jia et~al.}{2021}]{jia2021scaling}
\begin{bchapter}
\bauthor{\bsnm{Jia}, \binits{C.}},
\bauthor{\bsnm{Yang}, \binits{Y.}},
\bauthor{\bsnm{Xia}, \binits{Y.}},
\bauthor{\bsnm{Chen}, \binits{Y.-T.}},
\bauthor{\bsnm{Parekh}, \binits{Z.}},
\bauthor{\bsnm{Pham}, \binits{H.}},
\bauthor{\bsnm{Le}, \binits{Q.}},
\bauthor{\bsnm{Sung}, \binits{Y.-H.}},
\bauthor{\bsnm{Li}, \binits{Z.}},
\bauthor{\bsnm{Duerig}, \binits{T.}}:
\bctitle{Scaling up visual and vision-language representation learning with noisy text supervision}.
In: \bbtitle{International Conference on Machine Learning}
(\byear{2021})
\end{bchapter}
\endbibitem

\bibitem[\protect\citeauthoryear{Zhang et~al.}{2021}]{zhang2021vinvl}
\begin{bchapter}
\bauthor{\bsnm{Zhang}, \binits{P.}},
\bauthor{\bsnm{Li}, \binits{X.}},
\bauthor{\bsnm{Hu}, \binits{X.}},
\bauthor{\bsnm{Yang}, \binits{J.}},
\bauthor{\bsnm{Zhang}, \binits{L.}},
\bauthor{\bsnm{Wang}, \binits{L.}},
\bauthor{\bsnm{Choi}, \binits{Y.}},
\bauthor{\bsnm{Gao}, \binits{J.}}:
\bctitle{Vinvl: Revisiting visual representations in vision-language models}.
In: \bbtitle{Proceedings of the IEEE/CVF Conference on Computer Vision and Pattern Recognition},
pp. \bfpage{5579}--\blpage{5588}
(\byear{2021})
\end{bchapter}
\endbibitem

\bibitem[\protect\citeauthoryear{Chen et~al.}{2024}]{chen2024towards}
\begin{botherref}
\oauthor{\bsnm{Chen}, \binits{R.J.}},
\oauthor{\bsnm{Ding}, \binits{T.}},
\oauthor{\bsnm{Lu}, \binits{M.Y.}},
\oauthor{\bsnm{Williamson}, \binits{D.F.}},
\oauthor{\bsnm{Jaume}, \binits{G.}},
\oauthor{\bsnm{Song}, \binits{A.H.}},
\oauthor{\bsnm{Chen}, \binits{B.}},
\oauthor{\bsnm{Zhang}, \binits{A.}},
\oauthor{\bsnm{Shao}, \binits{D.}},
\oauthor{\bsnm{Shaban}, \binits{M.}}, et al.:
Towards a general-purpose foundation model for computational pathology.
Nature Medicine,
1--13
(2024)
\end{botherref}
\endbibitem

\bibitem[\protect\citeauthoryear{Lu et~al.}{2024}]{lu2024visual}
\begin{botherref}
\oauthor{\bsnm{Lu}, \binits{M.Y.}},
\oauthor{\bsnm{Chen}, \binits{B.}},
\oauthor{\bsnm{Williamson}, \binits{D.F.}},
\oauthor{\bsnm{Chen}, \binits{R.J.}},
\oauthor{\bsnm{Liang}, \binits{I.}},
\oauthor{\bsnm{Ding}, \binits{T.}},
\oauthor{\bsnm{Jaume}, \binits{G.}},
\oauthor{\bsnm{Odintsov}, \binits{I.}},
\oauthor{\bsnm{Le}, \binits{L.P.}},
\oauthor{\bsnm{Gerber}, \binits{G.}}, et al.:
A visual-language foundation model for computational pathology.
Nature Medicine,
1--12
(2024)
\end{botherref}
\endbibitem

\bibitem[\protect\citeauthoryear{Zhou et~al.}{2022a}]{zhou2022learning}
\begin{barticle}
\bauthor{\bsnm{Zhou}, \binits{K.}},
\bauthor{\bsnm{Yang}, \binits{J.}},
\bauthor{\bsnm{Loy}, \binits{C.C.}},
\bauthor{\bsnm{Liu}, \binits{Z.}}:
\batitle{Learning to prompt for vision-language models}.
\bjtitle{International Journal of Computer Vision}
\bvolume{130}(\bissue{9}),
\bfpage{2337}--\blpage{2348}
(\byear{2022})
\end{barticle}
\endbibitem

\bibitem[\protect\citeauthoryear{Zhou et~al.}{2022b}]{zhou2022conditional}
\begin{bchapter}
\bauthor{\bsnm{Zhou}, \binits{K.}},
\bauthor{\bsnm{Yang}, \binits{J.}},
\bauthor{\bsnm{Loy}, \binits{C.C.}},
\bauthor{\bsnm{Liu}, \binits{Z.}}:
\bctitle{Conditional prompt learning for vision-language models}.
In: \bbtitle{Proceedings of the IEEE/CVF Conference on Computer Vision and Pattern Recognition}
(\byear{2022})
\end{bchapter}
\endbibitem

\bibitem[\protect\citeauthoryear{Khattak et~al.}{2023}]{khattak2023maple}
\begin{bchapter}
\bauthor{\bsnm{Khattak}, \binits{M.U.}},
\bauthor{\bsnm{Rasheed}, \binits{H.}},
\bauthor{\bsnm{Maaz}, \binits{M.}},
\bauthor{\bsnm{Khan}, \binits{S.}},
\bauthor{\bsnm{Khan}, \binits{F.S.}}:
\bctitle{Maple: Multi-modal prompt learning}.
In: \bbtitle{Proceedings of the IEEE/CVF Conference on Computer Vision and Pattern Recognition}
(\byear{2023})
\end{bchapter}
\endbibitem

\bibitem[\protect\citeauthoryear{Wang et~al.}{2022}]{wang2022learning}
\begin{bchapter}
\bauthor{\bsnm{Wang}, \binits{Z.}},
\bauthor{\bsnm{Zhang}, \binits{Z.}},
\bauthor{\bsnm{Lee}, \binits{C.-Y.}},
\bauthor{\bsnm{Zhang}, \binits{H.}},
\bauthor{\bsnm{Sun}, \binits{R.}},
\bauthor{\bsnm{Ren}, \binits{X.}},
\bauthor{\bsnm{Su}, \binits{G.}},
\bauthor{\bsnm{Perot}, \binits{V.}},
\bauthor{\bsnm{Dy}, \binits{J.}},
\bauthor{\bsnm{Pfister}, \binits{T.}}:
\bctitle{Learning to prompt for continual learning}.
In: \bbtitle{Proceedings of the IEEE/CVF Conference on Computer Vision and Pattern Recognition}
(\byear{2022})
\end{bchapter}
\endbibitem

\bibitem[\protect\citeauthoryear{L{\"u}ddecke and Ecker}{2022}]{luddecke2022image}
\begin{bchapter}
\bauthor{\bsnm{L{\"u}ddecke}, \binits{T.}},
\bauthor{\bsnm{Ecker}, \binits{A.}}:
\bctitle{Image segmentation using text and image prompts}.
In: \bbtitle{Proceedings of the IEEE/CVF Conference on Computer Vision and Pattern Recognition},
pp. \bfpage{7086}--\blpage{7096}
(\byear{2022})
\end{bchapter}
\endbibitem

\bibitem[\protect\citeauthoryear{Rao et~al.}{2022}]{rao2022denseclip}
\begin{bchapter}
\bauthor{\bsnm{Rao}, \binits{Y.}},
\bauthor{\bsnm{Zhao}, \binits{W.}},
\bauthor{\bsnm{Chen}, \binits{G.}},
\bauthor{\bsnm{Tang}, \binits{Y.}},
\bauthor{\bsnm{Zhu}, \binits{Z.}},
\bauthor{\bsnm{Huang}, \binits{G.}},
\bauthor{\bsnm{Zhou}, \binits{J.}},
\bauthor{\bsnm{Lu}, \binits{J.}}:
\bctitle{Denseclip: Language-guided dense prediction with context-aware prompting}.
In: \bbtitle{Proceedings of the IEEE/CVF Conference on Computer Vision and Pattern Recognition}
(\byear{2022})
\end{bchapter}
\endbibitem

\bibitem[\protect\citeauthoryear{Ge et~al.}{2023}]{ge2023domain}
\begin{botherref}
\oauthor{\bsnm{Ge}, \binits{C.}},
\oauthor{\bsnm{Huang}, \binits{R.}},
\oauthor{\bsnm{Xie}, \binits{M.}},
\oauthor{\bsnm{Lai}, \binits{Z.}},
\oauthor{\bsnm{Song}, \binits{S.}},
\oauthor{\bsnm{Li}, \binits{S.}},
\oauthor{\bsnm{Huang}, \binits{G.}}:
Domain adaptation via prompt learning.
IEEE Transactions on Neural Networks and Learning Systems
(2023)
\end{botherref}
\endbibitem

\bibitem[\protect\citeauthoryear{Xu et~al.}{2024}]{xu2024worth}
\begin{botherref}
\oauthor{\bsnm{Xu}, \binits{H.}},
\oauthor{\bsnm{Chen}, \binits{Y.}},
\oauthor{\bsnm{Zhang}, \binits{D.}}:
Worth of prior knowledge for enhancing deep learning.
Nexus
(2024)
\end{botherref}
\endbibitem

\bibitem[\protect\citeauthoryear{He et~al.}{2022}]{he2022masked}
\begin{bchapter}
\bauthor{\bsnm{He}, \binits{K.}},
\bauthor{\bsnm{Chen}, \binits{X.}},
\bauthor{\bsnm{Xie}, \binits{S.}},
\bauthor{\bsnm{Li}, \binits{Y.}},
\bauthor{\bsnm{Doll{\'a}r}, \binits{P.}},
\bauthor{\bsnm{Girshick}, \binits{R.}}:
\bctitle{Masked autoencoders are scalable vision learners}.
In: \bbtitle{Proceedings of the IEEE/CVF Conference on Computer Vision and Pattern Recognition}
(\byear{2022})
\end{bchapter}
\endbibitem

\bibitem[\protect\citeauthoryear{Wang et~al.}{2022}]{wang2022advancing}
\begin{barticle}
\bauthor{\bsnm{Wang}, \binits{D.}},
\bauthor{\bsnm{Zhang}, \binits{Q.}},
\bauthor{\bsnm{Xu}, \binits{Y.}},
\bauthor{\bsnm{Zhang}, \binits{J.}},
\bauthor{\bsnm{Du}, \binits{B.}},
\bauthor{\bsnm{Tao}, \binits{D.}},
\bauthor{\bsnm{Zhang}, \binits{L.}}:
\batitle{Advancing plain vision transformer towards remote sensing foundation model}.
\bjtitle{IEEE Transactions on Geoscience and Remote Sensing}
\bvolume{61},
\bfpage{1}--\blpage{15}
(\byear{2022})
\end{barticle}
\endbibitem

\bibitem[\protect\citeauthoryear{Sun et~al.}{2022}]{sun2022ringmo}
\begin{botherref}
\oauthor{\bsnm{Sun}, \binits{X.}},
\oauthor{\bsnm{Wang}, \binits{P.}},
\oauthor{\bsnm{Lu}, \binits{W.}},
\oauthor{\bsnm{Zhu}, \binits{Z.}},
\oauthor{\bsnm{Lu}, \binits{X.}},
\oauthor{\bsnm{He}, \binits{Q.}},
\oauthor{\bsnm{Li}, \binits{J.}},
\oauthor{\bsnm{Rong}, \binits{X.}},
\oauthor{\bsnm{Yang}, \binits{Z.}},
\oauthor{\bsnm{Chang}, \binits{H.}}, et al.:
Ringmo: A remote sensing foundation model with masked image modeling.
IEEE Transactions on Geoscience and Remote Sensing
(2022)
\end{botherref}
\endbibitem

\bibitem[\protect\citeauthoryear{Ma and Wang}{2023}]{ma2023segment}
\begin{botherref}
\oauthor{\bsnm{Ma}, \binits{J.}},
\oauthor{\bsnm{Wang}, \binits{B.}}:
Segment anything in medical images.
arXiv:2304.12306
(2023)
\end{botherref}
\endbibitem

\bibitem[\protect\citeauthoryear{Parcollet et~al.}{2018}]{parcollet2018quaternion}
\begin{botherref}
\oauthor{\bsnm{Parcollet}, \binits{T.}},
\oauthor{\bsnm{Ravanelli}, \binits{M.}},
\oauthor{\bsnm{Morchid}, \binits{M.}},
\oauthor{\bsnm{Linar{\`e}s}, \binits{G.}},
\oauthor{\bsnm{Trabelsi}, \binits{C.}},
\oauthor{\bsnm{De~Mori}, \binits{R.}},
\oauthor{\bsnm{Bengio}, \binits{Y.}}:
Quaternion recurrent neural networks.
arXiv:1806.04418
(2018)
\end{botherref}
\endbibitem

\bibitem[\protect\citeauthoryear{Hu et~al.}{2021}]{hu2021lora}
\begin{botherref}
\oauthor{\bsnm{Hu}, \binits{E.J.}},
\oauthor{\bsnm{Shen}, \binits{Y.}},
\oauthor{\bsnm{Wallis}, \binits{P.}},
\oauthor{\bsnm{Allen-Zhu}, \binits{Z.}},
\oauthor{\bsnm{Li}, \binits{Y.}},
\oauthor{\bsnm{Wang}, \binits{S.}},
\oauthor{\bsnm{Wang}, \binits{L.}},
\oauthor{\bsnm{Chen}, \binits{W.}}:
Lora: Low-rank adaptation of large language models.
arXiv preprint arXiv:2106.09685
(2021)
\end{botherref}
\endbibitem

\bibitem[\protect\citeauthoryear{Qi et~al.}{2020}]{qi2020mlrsnet}
\begin{barticle}
\bauthor{\bsnm{Qi}, \binits{X.}},
\bauthor{\bsnm{Zhu}, \binits{P.}},
\bauthor{\bsnm{Wang}, \binits{Y.}},
\bauthor{\bsnm{Zhang}, \binits{L.}},
\bauthor{\bsnm{Peng}, \binits{J.}},
\bauthor{\bsnm{Wu}, \binits{M.}},
\bauthor{\bsnm{Chen}, \binits{J.}},
\bauthor{\bsnm{Zhao}, \binits{X.}},
\bauthor{\bsnm{Zang}, \binits{N.}},
\bauthor{\bsnm{Mathiopoulos}, \binits{P.T.}}:
\batitle{Mlrsnet: A multi-label high spatial resolution remote sensing dataset for semantic scene understanding}.
\bjtitle{ISPRS Journal of Photogrammetry and Remote Sensing}
\bvolume{169},
\bfpage{337}--\blpage{350}
(\byear{2020})
\end{barticle}
\endbibitem

\bibitem[\protect\citeauthoryear{Zhou et~al.}{2018}]{zhou2018patternnet}
\begin{barticle}
\bauthor{\bsnm{Zhou}, \binits{W.}},
\bauthor{\bsnm{Newsam}, \binits{S.}},
\bauthor{\bsnm{Li}, \binits{C.}},
\bauthor{\bsnm{Shao}, \binits{Z.}}:
\batitle{Patternnet: A benchmark dataset for performance evaluation of remote sensing image retrieval}.
\bjtitle{ISPRS journal of photogrammetry and remote sensing}
\bvolume{145},
\bfpage{197}--\blpage{209}
(\byear{2018})
\end{barticle}
\endbibitem

\bibitem[\protect\citeauthoryear{Zou et~al.}{2015}]{zou2015deep}
\begin{barticle}
\bauthor{\bsnm{Zou}, \binits{Q.}},
\bauthor{\bsnm{Ni}, \binits{L.}},
\bauthor{\bsnm{Zhang}, \binits{T.}},
\bauthor{\bsnm{Wang}, \binits{Q.}}:
\batitle{Deep learning based feature selection for remote sensing scene classification}.
\bjtitle{IEEE Geoscience and remote sensing letters}
\bvolume{12}(\bissue{11}),
\bfpage{2321}--\blpage{2325}
(\byear{2015})
\end{barticle}
\endbibitem

\bibitem[\protect\citeauthoryear{Xia et~al.}{2017}]{xia2017aid}
\begin{barticle}
\bauthor{\bsnm{Xia}, \binits{G.-S.}},
\bauthor{\bsnm{Hu}, \binits{J.}},
\bauthor{\bsnm{Hu}, \binits{F.}},
\bauthor{\bsnm{Shi}, \binits{B.}},
\bauthor{\bsnm{Bai}, \binits{X.}},
\bauthor{\bsnm{Zhong}, \binits{Y.}},
\bauthor{\bsnm{Zhang}, \binits{L.}},
\bauthor{\bsnm{Lu}, \binits{X.}}:
\batitle{Aid: A benchmark data set for performance evaluation of aerial scene classification}.
\bjtitle{IEEE Transactions on Geoscience and Remote Sensing}
\bvolume{55},
\bfpage{3965}--\blpage{3981}
(\byear{2017})
\end{barticle}
\endbibitem

\bibitem[\protect\citeauthoryear{Lu et~al.}{2017}]{lu2017exploring}
\begin{barticle}
\bauthor{\bsnm{Lu}, \binits{X.}},
\bauthor{\bsnm{Wang}, \binits{B.}},
\bauthor{\bsnm{Zheng}, \binits{X.}},
\bauthor{\bsnm{Li}, \binits{X.}}:
\batitle{Exploring models and data for remote sensing image caption generation}.
\bjtitle{IEEE Transactions on Geoscience and Remote Sensing}
\bvolume{56}(\bissue{4}),
\bfpage{2183}--\blpage{2195}
(\byear{2017})
\end{barticle}
\endbibitem

\bibitem[\protect\citeauthoryear{Yang and Newsam}{2010}]{yang2010bag}
\begin{bchapter}
\bauthor{\bsnm{Yang}, \binits{Y.}},
\bauthor{\bsnm{Newsam}, \binits{S.}}:
\bctitle{Bag-of-visual-words and spatial extensions for land-use classification}.
In: \bbtitle{Proceedings of the 18th SIGSPATIAL International Conference on Advances in Geographic Information Systems}
(\byear{2010})
\end{bchapter}
\endbibitem

\bibitem[\protect\citeauthoryear{Dai and Yang}{2011}]{Dai2011WHURS19}
\begin{barticle}
\bauthor{\bsnm{Dai}, \binits{D.}},
\bauthor{\bsnm{Yang}, \binits{W.}}:
\batitle{Satellite image classification via two-layer sparse coding with biased image representation}.
\bjtitle{IEEE Transactions on Geoscience and Remote Sensing}
\bvolume{8}(\bissue{1}),
\bfpage{173}--\blpage{176}
(\byear{2011})
\end{barticle}
\endbibitem

\bibitem[\protect\citeauthoryear{Cheng et~al.}{2017}]{cheng2017remote}
\begin{barticle}
\bauthor{\bsnm{Cheng}, \binits{G.}},
\bauthor{\bsnm{Han}, \binits{J.}},
\bauthor{\bsnm{Lu}, \binits{X.}}:
\batitle{Remote sensing image scene classification: Benchmark and state of the art}.
\bjtitle{Proceedings of the IEEE}
\bvolume{105}(\bissue{10}),
\bfpage{1865}--\blpage{1883}
(\byear{2017})
\end{barticle}
\endbibitem

\bibitem[\protect\citeauthoryear{Nickparvar}{2021}]{BTMRI}
\begin{botherref}
\oauthor{\bsnm{Nickparvar}, \binits{M.}}:
Brain Tumor MRI Dataset.
Kaggle
(2021).
\doiurl{10.34740/KAGGLE/DSV/2645886} .
\url{https://www.kaggle.com/dsv/2645886}
\end{botherref}
\endbibitem

\bibitem[\protect\citeauthoryear{Hashemi}{2023}]{CCBTM}
\begin{botherref}
\oauthor{\bsnm{Hashemi}, \binits{S.M.H.}}:
Crystal Clean: Brain Tumors MRI Dataset.
Kaggle
(2023).
\doiurl{10.34740/KAGGLE/DS/3505991} .
\url{https://www.kaggle.com/ds/3505991}
\end{botherref}
\endbibitem

\bibitem[\protect\citeauthoryear{Kather et~al.}{2016}]{CHMNIST}
\begin{botherref}
\oauthor{\bsnm{Kather}, \binits{J.}},
\oauthor{\bsnm{Weis}, \binits{C.}},
\oauthor{\bsnm{Bianconi}, \binits{F.}},
\oauthor{\bsnm{Melchers}, \binits{S.}},
\oauthor{\bsnm{Schad}, \binits{L.}},
\oauthor{\bsnm{Gaiser}, \binits{T.}},
\oauthor{\bsnm{Marx}, \binits{A.}},
\oauthor{\bsnm{F}, \binits{Z.}}:
Multi-class texture analysis in colorectal cancer histology.
Scientific Reports
(2016)
\end{botherref}
\endbibitem

\bibitem[\protect\citeauthoryear{Neumann et~al.}{2020}]{neumann202sandstones}
\begin{botherref}
\oauthor{\bsnm{Neumann}, \binits{R.}},
\oauthor{\bsnm{Andreeta}, \binits{M.}},
\oauthor{\bsnm{Lucas-Oliveira}, \binits{E.}}:
Sandstones: raw, filtered and segmented data. Digital Rocks Portal.
Online
(2020).
\url{http://www.digitalrocksportal.org/projects/317.}
\end{botherref}
\endbibitem

\bibitem[\protect\citeauthoryear{Moon and Andrew}{2019}]{moon2019intergranular}
\begin{botherref}
\oauthor{\bsnm{Moon}, \binits{C.}},
\oauthor{\bsnm{Andrew}, \binits{M.}}:
Intergranular pore structures in sandstones. Digital Rocks Portal.
Online
(2019).
\url{https://www.digitalrocksportal.org/projects/222.}
\end{botherref}
\endbibitem

\bibitem[\protect\citeauthoryear{Muljadi}{2015}]{muljadi2015estaillades}
\begin{botherref}
\oauthor{\bsnm{Muljadi}, \binits{B.}}:
Estaillades carbonate
(2015)
\end{botherref}
\endbibitem

\bibitem[\protect\citeauthoryear{Raeini et~al.}{2017}]{raeini2017generalized}
\begin{barticle}
\bauthor{\bsnm{Raeini}, \binits{A.Q.}},
\bauthor{\bsnm{Bijeljic}, \binits{B.}},
\bauthor{\bsnm{Blunt}, \binits{M.J.}}:
\batitle{Generalized network modeling: Network extraction as a coarse-scale discretization of the void space of porous media}.
\bjtitle{Physical Review E}
\bvolume{96}(\bissue{1}),
\bfpage{013312}
(\byear{2017})
\end{barticle}
\endbibitem

\bibitem[\protect\citeauthoryear{Mohammadmoradi}{2017}]{mohammadmoradi2017multiscale}
\begin{botherref}
\oauthor{\bsnm{Mohammadmoradi}, \binits{P.}}:
A multiscale sandy microstructure
(2017)
\end{botherref}
\endbibitem

\bibitem[\protect\citeauthoryear{national laboratory}{2014}]{MSTAR}
\begin{botherref}
\oauthor{\bsnm{laboratory}, \binits{S.}}:
The Air Force Moving and Stationary Target Recognition Database.
Online
(2014).
\url{https://www.sdms.afrl.af.mil/index.}
\end{botherref}
\endbibitem

\bibitem[\protect\citeauthoryear{Hou et~al.}{2020}]{hou2020fusar}
\begin{barticle}
\bauthor{\bsnm{Hou}, \binits{X.}},
\bauthor{\bsnm{Ao}, \binits{W.}},
\bauthor{\bsnm{Song}, \binits{Q.}},
\bauthor{\bsnm{Lai}, \binits{J.}},
\bauthor{\bsnm{Wang}, \binits{H.}},
\bauthor{\bsnm{Xu}, \binits{F.}}:
\batitle{Fusar-ship: Building a high-resolution sar-ais matchup dataset of gaofen-3 for ship detection and recognition}.
\bjtitle{Science China Information Sciences}
\bvolume{63},
\bfpage{1}--\blpage{19}
(\byear{2020})
\end{barticle}
\endbibitem

\bibitem[\protect\citeauthoryear{Sun et~al.}{2022}]{SAR-ACD}
\begin{barticle}
\bauthor{\bsnm{Sun}, \binits{X.}},
\bauthor{\bsnm{Lv}, \binits{Y.}},
\bauthor{\bsnm{Wang}, \binits{Z.}},
\bauthor{\bsnm{Fu}, \binits{K.}}:
\batitle{Scan: Scattering characteristics analysis network for few-shot aircraft classification in high-resolution sar images}.
\bjtitle{IEEE Transactions on Geoscience and Remote Sensing}
\bvolume{60},
\bfpage{1}--\blpage{17}
(\byear{2022})
\doiurl{10.1109/TGRS.2022.3166174}
\end{barticle}
\endbibitem

\bibitem[\protect\citeauthoryear{Dosovitskiy et~al.}{2020}]{dosovitskiy2020image}
\begin{botherref}
\oauthor{\bsnm{Dosovitskiy}, \binits{A.}},
\oauthor{\bsnm{Beyer}, \binits{L.}},
\oauthor{\bsnm{Kolesnikov}, \binits{A.}},
\oauthor{\bsnm{Weissenborn}, \binits{D.}},
\oauthor{\bsnm{Zhai}, \binits{X.}},
\oauthor{\bsnm{Unterthiner}, \binits{T.}},
\oauthor{\bsnm{Dehghani}, \binits{M.}},
\oauthor{\bsnm{Minderer}, \binits{M.}},
\oauthor{\bsnm{Heigold}, \binits{G.}},
\oauthor{\bsnm{Gelly}, \binits{S.}}, et al.:
An image is worth 16x16 words: Transformers for image recognition at scale.
arXiv:2010.11929
(2020)
\end{botherref}
\endbibitem

\bibitem[\protect\citeauthoryear{Xu et~al.}{2021}]{xu2021vitae}
\begin{botherref}
\oauthor{\bsnm{Xu}, \binits{Y.}},
\oauthor{\bsnm{Zhang}, \binits{Q.}},
\oauthor{\bsnm{Zhang}, \binits{J.}},
\oauthor{\bsnm{Tao}, \binits{D.}}:
Vitae: Vision transformer advanced by exploring intrinsic inductive bias.
Advances in Neural Information Processing Systems
(2021)
\end{botherref}
\endbibitem

\bibitem[\protect\citeauthoryear{Xiaoyan}{2023}]{SAR-OOD}
\begin{botherref}
\oauthor{\bsnm{Xiaoyan}, \binits{Z.}}:
SAR-OOD-Detection-Data.
Online
(2023).
\url{https://github.com/Xiaoyan-Zhou/SAR-OOD-Detection-Data/tree/main}
\end{botherref}
\endbibitem

\end{thebibliography}

\clearpage
\appendix
\begin{center}
\huge
 Supplemental Materials
\end{center}

\section{Dataset Details}
To testify the performance of our proposed generalized domain prompt learning,  five distinctive domains including remote sensing, medical imaging, geology, SAR, and fluid dynamics are adopted for experiments, respectively providing domain-specific VLMs for corresponding domain researchers. Particularly, for the remote sensing domain, eight remote sensing datasets, namely MLRSNet~\cite{qi2020mlrsnet}, PatternNet~\cite{zhou2018patternnet}, RSSCN7~\cite{zou2015deep}, AID~\cite{xia2017aid}, RSICD~\cite{lu2017exploring}, UCM~\cite{yang2010bag}, WHURS19~\cite{Dai2011WHURS19}, and NWPU~\cite{cheng2017remote}, are leveraged to conduct the experiments.  MLRSNet dataset consists of 46 categories with 109,161 annotated remote sensing images, each having a size of 224 × 224 × 3 pixels. PatternNet dataset comprises 38 categories, with each class containing 800 images of size 256 × 256 × 3 pixels. RSSCN7 dataset is relatively smaller, comprising only 7 categories, with each category containing 400 images of size 400 × 400 × 3 pixels. AID dataset consists of 30 classes and a total of 10,000 images, with each image having a size of 600 × 600 × 3 pixels. The number of images per class in AID varies. RSICD dataset contains 10,921 images with varying resolutions, belonging to 30 different categories. UCM dataset consists of 21 scene classes, with 100 samples of size 256 × 256 × 3 pixels in each category. WHURS19 dataset comprises 19 different scene classes, with 50 samples of size 600 × 600 × 3 pixels in each class. NWPU dataset consists of 40 categories, with each class containing 700 images of size 256 × 256 × 3 pixels.

To testify the constructed medical VLMs, we adopt three medical image recognition datasets for experiments: the Brain Tumor MRI Dataset (BTMRI)~\cite{BTMRI}, Colorectal Histology MNIST (CHMNIST)~\cite{CHMNIST}, and Crystal Clean: Brain Tumors MRI Dataset (CCBTM)~\cite{CCBTM}. The BTMRI dataset consists of 7023 human brain MRI images categorized into four classes: glioma, meningioma, no-tumor, and pituitary. Similarly, the CCBTM dataset focuses on brain tumor classification, comprising 3264 MRI images across four categories. The CHMNIST dataset is a tissue classification dataset that focuses on histology tiles from patients with colorectal cancer, containing eight categories.

To evaluate the obtained SAR VLMs, We leveraged the famous MSTAR~\cite{MSTAR}, FUSAR-ship~\cite{hou2020fusar}, and SAR-ACD~\cite{SAR-ACD} as the evaluation benchmark to testify our proposed method. The Moving and Stationary Target Acquisition and Recognition (MSTAR) dataset was collected and proposed by Sandia National Laboratory, it consists of SAR slice images of varying stationary vehicles with a pixel size of 128×128. FUSRA-ship is a SAR ship classification dataset, which contains 8 different ship categories and each category has about 500 samples.  The SAR-ACD datasets provide 6 civil aircraft categories for classification. These images of HH polarization in different time phases cover the Shanghai Hongqiao Airport area and other airports.  This public dataset contains 3032 images with around 500 images in each category.

In the realm of geology, we adopt the recognition of digital rocks as representative instances, offering basic rock VLMs for researchers. Specifically, we procure five types of digital rock images from publicly available datasets: Berea sandstone~\cite{neumann202sandstones}, Doddington sandstone~\cite{moon2019intergranular}, Estaillade carbonate~\cite{muljadi2015estaillades}, Ketton carbonate~\cite{raeini2017generalized}, and Sandy multiscale medium ~\cite{mohammadmoradi2017multiscale}. Although these original rock images are three-dimensional shapes with dimensions of $64^3$, experts in rock analysis can discern the rock types directly from their slices. Taking inspiration from this approach,  we opt to utilize the central slices along the XYZ axes of the rocks for the purposes of our experimental investigations

Aiming at building and evaluating fluids dynamics VLMs, the utilized flow images dataset comes from the direct numerical simulation (DNS) data with the commonly used Navier-Stokes (NS) equations. Particularly, the flow images are acquired from simulating incompressible turbulence undergoing natural decay at a  Reynolds number of approximately $Re_{\lambda} \approx 250$. The simulations were conducted on a uniform grid resolution of $512^3$, and we utilized the flow slice in XYZ dimensions or UVW dimensions to conduct the experiments. In evaluation, the model should accurately classify the 512×512 flow slice images into the corresponding dimensions. The simulation data is collected every 2000 steps within the range of 30000 to 50000 steps, resulting in 10 different sub-step datasets. For evaluation purposes, we respectively utilize the datasets from the 30000-step, 40000-step, and 50000-step time points.

\section{Quaternion Networks}
In the four-dimensional space, a quaternion $Q$ extends beyond a complex number and can be denoted as:
\begin{equation}
 Q= r1+ x\mathbf{i} + y\mathbf{j} + z\mathbf{k},
\end{equation}
where $r$, $x$, $y$, and $z$ are real numbers, and $1$, $\mathbf{i}$, $\mathbf{j}$, and $\mathbf{k}$ are the quaternion unit basis. Here, $r$ represents the real part of $Q$, while $x\mathbf{i} + y\mathbf{j} + z\mathbf{k}$ denotes its imaginary or vector part. Quaternions find utility in describing spatial rotations and other applications due to their embedded information, which can be represented by a real-number matrix:
\begin{equation}
{Q} = \left[ {\begin{array}{*{20}{c}}
r&{ - x}&{ - y}&{ - z}\\
x&r&{ - z}&y\\
y&z&r&{ - x}\\
z&{ - y}&x&r
\end{array}} \right].
\end{equation}

A quaternion neural network is defined as:
\begin{equation}
{Q_{\text{out}}} = \alpha ( W \otimes Q),
\end{equation}
where $W$ represents the learnable parameters of the quaternion neural networks, $\otimes$ denotes the Hadamard product, and $\alpha$ is the activation function defined as:
\begin{equation}
\alpha (Q) = f(r)1 + f(x)\mathbf{i} + f(y)\mathbf{j} + f(z)\mathbf{k},
\end{equation}
where $f$ is any standard activation function. The Hamilton product $\otimes$ of two quaternions $Q_1$ and $Q_2$ is computed as:
\begin{equation}
\begin{split}
{Q_1} \otimes {Q_2} = & ({r_1}{r_2} - {x_1}{x_2} - {y_1}{y_2} - {z_1}{z_2}) \\ & + ({r_1}{x_2} + 
 {x_1}{r_2} + {y_1}{z_2} - {z_1}{y_2})\textbf{i} \\ & + ({r_1}{y_2} - {x_1}{z_2} +
 {y_1}{r_2} + {z_1}{x_2})\textbf{j} \\ & + ({r_1}{z_2} + {x_1}{y_2} - {y_1}{x_2} + {z_1}{r_2})\textbf{k} ,
\end{split}
\end{equation}

We propose utilizing quaternion networks to explore orthogonal intermodal relationships between domain-specific vision features from the domain-specific foundation models and contextual embeddings from the language branch, inspired by the distinctive feature processing pattern of quaternion networks.

\begin{table*}[t!]
\small
\tablestyle{5pt}{0}
\addtolength{\tabcolsep}{-6pt}
    \tabstyle{1.0pt}
    \setlength{\tabcolsep}{1pt}
    \caption{Comparison with SOTA methods in base-to-novel generalization on 8 remote sensing recognition datasets. Our method consistently performs well over the SOTA approaches. We use \red{red} and \blu{blue} to highlight the first and second best scores. Except the HM denotes the harmonic mean score, others are the accuracy scores. }
    \scalebox{0.75}{
    \begin{subtable}[t]{.32\textwidth}
    \centering
    \caption{\textbf{Average over 8 datasets}}
    \begin{tabular}{l cc c}
    \toprule
    & Base & Novel & HM \\
    \midrule
    CLIP~\cite{radford2021learning} & 71.19	&71.33	&70.63 \\
    CoOp~\cite{zhou2022learning} & 87.61	&70.84	&78.03 \\
    CoCoOp~\cite{zhou2022conditional} & 91.82	&68.98	&78.43 \\
    MaPLe~\cite{khattak2023maple} & 93.12	&71.71	&80.42 \\
    \midrule
    Ours (ViTAE)   & \red{94.53} &\blu{74.45}	&\blu{82.86}
\\
    \rowcolor{tabhighlight}
       Ours (ViT) &\blu{94.46}	&\red{75.27}	&\red{84.67}
\\
    \bottomrule
    \end{tabular}
    \end{subtable}
    }
        \scalebox{0.75}{
    \begin{subtable}[t]{.32\textwidth}
    \centering
    \caption{MLRSNet}
    \begin{tabular}{l cc c}
    \toprule
    & Base & Novel & HM \\
    \midrule
    CLIP~\cite{radford2021learning} & 64.50 & 60.30 & 62.33 \\
    CoOp~\cite{zhou2022learning} & 79.37 & 58.90 & 67.62\\
    CoCoOp~\cite{zhou2022conditional} & 83.30 & 59.50 & 69.42 \\
    MaPLe~\cite{khattak2023maple} & 85.23 & 59.60 & 70.15 \\
    \midrule
    Ours (ViTAE)  &\red{89.24}	&\red{60.63}	&\red{72.20}
\\
    \rowcolor{tabhighlight}
     Ours (ViT) & \blu{89.00}	&\blu{60.37}	&\blu{71.94}  \\
    \bottomrule
    \end{tabular}
    \end{subtable}
    }
    ~
    \scalebox{0.75}{
    \begin{subtable}[t]{.32\textwidth}
    \centering
    \caption{PatternNet}
    \begin{tabular}{l cc c}
    \toprule
    & Base & Novel & HM \\
    \midrule
    CLIP~\cite{radford2021learning} & 70.60 & 62.60 & 66.36 \\
    CoOp~\cite{zhou2022learning} & 87.30 & \blu{64.20} & 73.99 \\
    CoCoOp~\cite{zhou2022conditional} & 93.70 & 59.90 & 73.08 \\
    MaPLe~\cite{khattak2023maple} & \blu{95.30} & 57.90 & 72.03 \\
    \midrule
    Ours (ViTAE) & \red{96.93}	&63.80	& \blu{76.95} \\
    \rowcolor{tabhighlight}
     Ours (ViT) & \red{96.93}	& \red{64.30}	& \red{77.31} \\
    \bottomrule
    \end{tabular}
    \end{subtable}
    }
    ~
        \scalebox{0.75}{
    \begin{subtable}[t]{.32\textwidth}
    \centering
    \caption{RSSCN7}
    \begin{tabular}{l cc c}
    \toprule
    & Base & Novel & HM \\
    \midrule
    CLIP~\cite{radford2021learning} & 66.70 & 95.30 & 78.48 \\
    CoOp~\cite{zhou2022learning} & 84.80  &	89.13 & 86.91 \\
    CoCoOp~\cite{zhou2022conditional} & 90.97	&90.00	&90.48 \\
    MaPLe~\cite{khattak2023maple} & \blu{91.67}	&\blu{93.70}	&\blu{92.67} \\
    \midrule
 Ours (ViTAE) &\red{91.90}	&92.80	&92.35
\\
    \rowcolor{tabhighlight}
    Ours (ViT) &91.07	&\red{95.13}	&\red{93.06}\\
    \bottomrule
    \end{tabular}
    \end{subtable}
        }
        \scalebox{0.75}{
    \begin{subtable}[t]{.32\textwidth}
    \centering
    \caption{AID}
    \begin{tabular}{l cc c}
    \toprule
    & Base & Novel & HM \\
    \midrule
    CLIP~\cite{radford2021learning} & 73.50	&70.40 &71.92 \\
    CoOp~\cite{zhou2022learning} & 87.63	&70.37	&78.06\\
    CoCoOp~\cite{zhou2022conditional} &92.63	&65.73	&76.89 \\
    MaPLe~\cite{khattak2023maple} & 92.73	&\blu{74.57}	&82.66 \\
    \midrule
    Ours (ViTAE) &\blu{93.87}	&73.93	& \blu{82.72}
 \\
    \rowcolor{tabhighlight}
   Ours (ViT) & \red{94.47}	& \red{75.43}	&\red{83.88} \\
    \bottomrule
    \end{tabular}
    \end{subtable}
    }
    ~
        \scalebox{0.75}{
    \begin{subtable}[t]{.32\textwidth}
    \centering
    \caption{RSICD}
    \begin{tabular}{l cc c}
    \toprule
    & Base & Novel & HM \\
    \midrule
    CLIP~\cite{radford2021learning} & 71.50   &60.20	&65.37 \\
    CoOp~\cite{zhou2022learning} & 88.43	&60.20	&71.63 \\
    CoCoOp~\cite{zhou2022conditional} & 92.37	&58.80	&71.86 \\
    MaPLe~\cite{khattak2023maple} & 93.93	&56.27	&70.38 \\
    \midrule
    Ours (ViTAE) &\blu{95.33}	 &\blu{66.10}	&\blu{78.07}\\
    \rowcolor{tabhighlight}
    Ours (ViT) & \red{95.50}	& \red{66.95}	&\red{78.72} \\
    \bottomrule
    \end{tabular}
    \end{subtable}
    }
    ~
        \scalebox{0.75}{
    \begin{subtable}[t]{.32\textwidth}
    \centering
    \caption{UCM}
    \begin{tabular}{l cc c}
    \toprule
    & Base & Novel & HM \\
    \midrule
    CLIP~\cite{radford2021learning} & 80.60	&68.00	&73.77 \\
    CoOp~\cite{zhou2022learning} & 93.60	&\blu{74.53}	&\blu{82.98} \\
    CoCoOp~\cite{zhou2022conditional}& 95.23	&71.57	&81.72 \\
    MaPLe~\cite{khattak2023maple} & \blu{97.70}	&70.90	&82.17 \\
    \midrule
    Ours (ViTAE) &\blu{97.70} 	&72.10	&82.97\\
    \rowcolor{tabhighlight}
    Ours (ViT) &\red{97.90}	&\red{75.23}	&\red{85.08}\\
    \bottomrule
    \end{tabular}
    \end{subtable}
    }
        \scalebox{0.75}{
    \begin{subtable}[t]{.32\textwidth}
    \centering
    \caption{WHURS19}
    \begin{tabular}{l cc c}
    \toprule
    & Base & Novel & HM \\
    \midrule
    CLIP~\cite{radford2021learning} & 73.10	&\blu{90.80}  &80.99 \\
    CoOp~\cite{zhou2022learning} & 95.20	 &82.40	&88.34 \\
    CoCoOp~\cite{zhou2022conditional} & 97.10	&77.00	&85.89 \\
    MaPLe~\cite{khattak2023maple} & 97.70	&88.03	&92.61 \\
    \midrule
    Ours (ViTAE) &\red{99.20}	&\red{92.27}	&\red{95.61}\\
    \rowcolor{tabhighlight}
    Ours (ViT)  &\blu{98.63}	&89.90	&\blu{94.06}\\
    \bottomrule
    \end{tabular}
    \end{subtable}
    }
    ~
        \scalebox{0.75}{
    \begin{subtable}[t]{.32\textwidth}
    \centering
    \caption{NWPU}
    \begin{tabular}{l cc c}
    \toprule
    & Base & Novel & HM \\
    \midrule
    CLIP~\cite{radford2021learning} & 69.00	&63.00	&65.87 \\
    CoOp~\cite{zhou2022learning} & 84.53	&66.97	&74.73 \\
    CoCoOp~\cite{zhou2022conditional} & 89.27	&69.37	&78.07 \\
    MaPLe~\cite{khattak2023maple} & 90.70	&72.70	&80.71\\
    \midrule
    Ours (ViTAE) &\blu{92.03}	 &\blu{73.95}	&\blu{82.01}\\
    \rowcolor{tabhighlight}
     Ours (ViT) &\red{92.20}	&\red{74.83}	&\red{82.61} \\
    \bottomrule
    \end{tabular}
    \end{subtable}
    }

    \label{table1}
\end{table*}

\begin{table*}[t!]
\small
\tablestyle{6pt}{0}
\addtolength{\tabcolsep}{-6pt}
    \tabstyle{1.0pt}
    \setlength{\tabcolsep}{1pt}
    \caption{Comparison between our method and SOTA methods for base-to-novel generalization on medical image classification datasets. Our method performs well over the compared methods. We use \red{red} and \blu{blue} to indicate the first and second best scores.  Except the HM denotes the harmonic mean score, others are the accuracy scores. 
    }
    \scalebox{0.7}{
    \begin{subtable}[t]{.25\textwidth}
    \centering
    \caption{\textbf{Average}}
        \begin{tabular}{l cc c}
    \toprule
    & Base & Novel & HM \\
    \midrule
    CLIP~\cite{radford2021learning} &49.83	&41.83	&45.18 \\
    CoOp~\cite{zhou2022learning} & 51.59	&43.77	&46.81 \\
    CoCoOp~\cite{zhou2022conditional}  & \blu{64.45}	&43.16	&\blu{49.45} \\
    MaPLe~\cite{khattak2023maple}  & 62.39	&\red{44.40} &49.01 \\
    \midrule
    \rowcolor{tabhighlight}
       Ours  &  \red{80.27}	&\blu{44.11}	&\red{54.64}\\
    \bottomrule
    \end{tabular}

    \end{subtable}
    }
        ~
    \begin{subtable}[t]{.25\textwidth}
    \centering
    \caption{All datasets}
\includegraphics[width=0.72\linewidth]{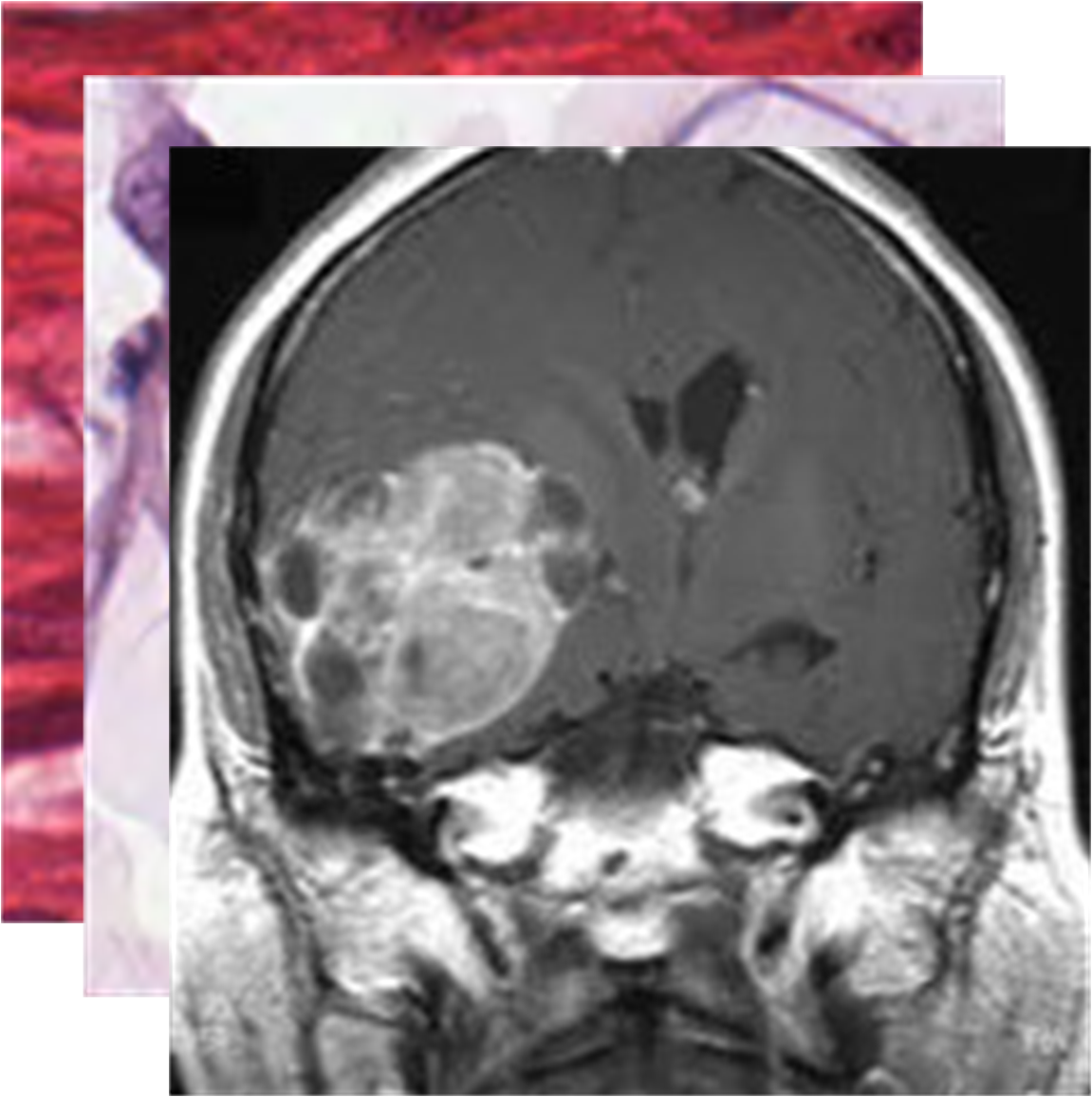}
    \end{subtable}
    ~
\scalebox{0.7}{
    \begin{subtable}[t]{.25\textwidth}
    \centering
    \caption{BTMRI}
    \begin{tabular}{l cc c}
    \toprule
    & Base & Novel & HM \\
    \midrule
    CLIP~\cite{radford2021learning}  &50.60	&51.20	&50.89 \\
    CoOp~\cite{zhou2022learning} & 48.93	&53.30	&51.02 \\
    CoCoOp~\cite{zhou2022conditional} & 52.37	&52.80	&52.58 \\
    MaPLe~\cite{khattak2023maple} & \blu{53.67}	&\red{61.60} &\blu{57.36} \\
    \midrule
    \rowcolor{tabhighlight}
       Ours  &\red{63.37}	&\blu{54.20}	&\red{58.43}\\
    \bottomrule
    \end{tabular}
    \end{subtable}
    }
            ~
    \begin{subtable}[t]{.25\textwidth}
    \centering
    \caption{BTMRI dataset}
\includegraphics[width=0.7\linewidth]{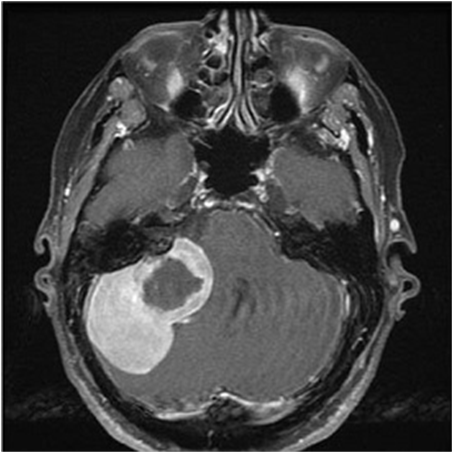}
    \end{subtable}
         ~
        \scalebox{0.7}{
    \begin{subtable}[t]{.25\textwidth}
    \centering
    \caption{CHMNIST}
    \begin{tabular}{l cc c}
    \toprule
    & Base & Novel & HM \\
    \midrule
    CLIP~\cite{radford2021learning} & 31.60 & \red{27.40} & 29.35 \\
    CoOp~\cite{zhou2022learning} & 41.70 & 25.67 & 31.78 \\
    CoCoOp~\cite{zhou2022conditional} & \blu{74.30} & 25.30 & \blu{37.74} \\
    MaPLe~\cite{khattak2023maple} & 74.03 & 25.10 & 37.49 \\
    \midrule
    \rowcolor{tabhighlight}
     Ours  & \red{92.45}	&\blu{26.67}	&\red{41.40}
 \\
    \bottomrule
    \end{tabular}
    \end{subtable}
    }
    ~
    \begin{subtable}[t]{.25\textwidth}
    \centering
    \caption{CHMNIST dataset}
\includegraphics[width=0.7\linewidth]{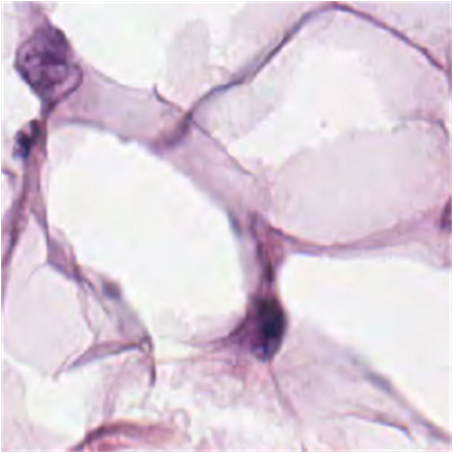}
    \end{subtable}
    ~
        \scalebox{0.7}{
    \begin{subtable}[t]{.25\textwidth}
    \centering
    \caption{CCBTM}
    \begin{tabular}{l cc c}
    \toprule
    & Base & Novel & HM \\
    \midrule
    CLIP~\cite{radford2021learning} & \blu{67.30} & 46.90 & 55.28 \\
    CoOp~\cite{zhou2022learning} & 64.13  &	\red{52.33} & 57.63 \\
    CoCoOp~\cite{zhou2022conditional} & 66.67	&51.37	&\blu{58.03} \\
    MaPLe~\cite{khattak2023maple} & 59.47	&46.50	&52.19 \\
    \midrule
    \rowcolor{tabhighlight}
    Ours  & \red{85.00}	&\blu{51.47}	&\red{64.12}\\
    \bottomrule
    \end{tabular}
    \end{subtable}
        }
        ~
    \begin{subtable}[t]{.25\textwidth}
    \centering
    \caption{CCBTM dataset}
\includegraphics[width=0.7\linewidth]{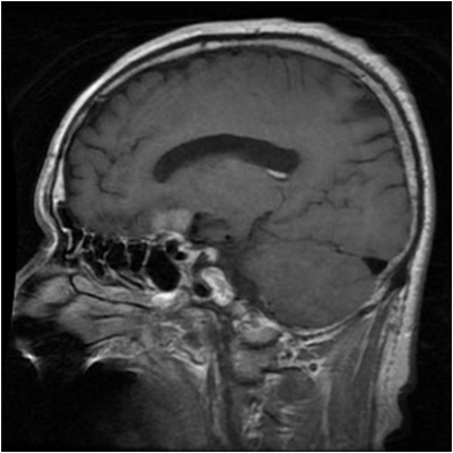}
    \end{subtable}
    \label{table2}
\end{table*}

\begin{table*}[t!]
\small
\tablestyle{6pt}{0}
\addtolength{\tabcolsep}{-6pt}
    \tabstyle{1.0pt}
    \setlength{\tabcolsep}{1pt}
    \caption{Comparison between our method and SOTA methods for base-to-novel generalization on rock image recognition datasets. Our method performs well over the compared methods. We use \red{red} and \blu{blue} to indicate the first and second-best scores.  Except the HM denotes the harmonic mean score, others are the accuracy scores.
    }
    \scalebox{0.7}{
    \begin{subtable}[t]{.25\textwidth}
    \centering
    \caption{\textbf{Average}}
        \begin{tabular}{l cc c}
    \toprule
    & Base & Novel & HM \\
    \midrule
    CLIP~\cite{radford2021learning} &32.53	&46.60	&38.31
 \\
    CoOp~\cite{zhou2022learning} & 27.07	&25.97	&26.49 \\
    CoCoOp~\cite{zhou2022conditional}  & \blu{82.35}	&50.42	&62.49 \\
    MaPLe~\cite{khattak2023maple}  & 78.97	&\blu{52.35}	&\blu{62.94} \\
    \midrule
    \rowcolor{tabhighlight}
       Ours  & \red{94.76}	&\red{57.40}	&\red{71.42}\\
    \bottomrule
    \end{tabular}

    \end{subtable}
    }
        ~
    \begin{subtable}[t]{.25\textwidth}
    \centering
    \caption{All datasets}
\includegraphics[width=0.72\linewidth]{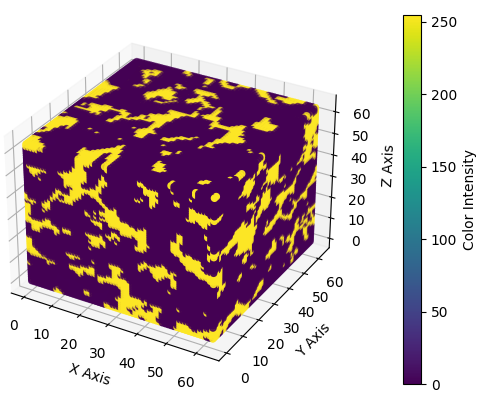}
    \end{subtable}
    ~
\scalebox{0.7}{
    \begin{subtable}[t]{.25\textwidth}
    \centering
    \caption{Rock X}
    \begin{tabular}{l cc c}
    \toprule
    & Base & Novel & HM \\
    \midrule
    CLIP~\cite{radford2021learning} & 31.10	&46.60	&37.30\\
    CoOp~\cite{zhou2022learning} & 26.00	&26.30	&26.15 \\
    CoCoOp~\cite{zhou2022conditional} & 80.70	&46.97	&59.38 \\
    MaPLe~\cite{khattak2023maple} & \blu{86.60}	&\red{54.70}	&\blu{67.05} \\
    \midrule
    \rowcolor{tabhighlight}
     Ours  & \red{90.73}	&\blu{54.27}	&\red{67.92}\\
    \bottomrule
    \end{tabular}
    \end{subtable}
    }
            ~
    \begin{subtable}[t]{.25\textwidth}
    \centering
    \caption{Rock X dataset}
\includegraphics[width=0.7\linewidth]{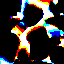}
    \end{subtable}
         ~
        \scalebox{0.7}{
    \begin{subtable}[t]{.25\textwidth}
    \centering
    \caption{Rock Y}
    \begin{tabular}{l cc c}
    \toprule
    & Base & Novel & HM \\
    \midrule
    CLIP~\cite{radford2021learning} & 33.80	&46.60	&39.18\\
    CoOp~\cite{zhou2022learning} &28.20	&25.50	&26.78 \\
    CoCoOp~\cite{zhou2022conditional} &\blu{88.33}	&\blu{52.27}	&\blu{65.68} \\
    MaPLe~\cite{khattak2023maple} &71.07	&48.00	&57.30 \\
    \midrule
    \rowcolor{tabhighlight}
     Ours  &\red{95.83}	&\red{63.10}	&\red{76.09}
 \\
    \bottomrule
    \end{tabular}
    \end{subtable}
    }
    ~
    \begin{subtable}[t]{.25\textwidth}
    \centering
    \caption{Rock Y dataset}
\includegraphics[width=0.7\linewidth]{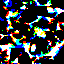}
    \end{subtable}
    ~
        \scalebox{0.7}{
    \begin{subtable}[t]{.25\textwidth}
    \centering
    \caption{Rock Z}
    \begin{tabular}{l cc c}
    \toprule
    & Base & Novel & HM \\
    \midrule
    CLIP~\cite{radford2021learning} &32.70	&46.60	&38.43 \\
    CoOp~\cite{zhou2022learning} & 27.00	&26.10	&26.54 \\
    CoCoOp~\cite{zhou2022conditional} &78.03	&52.03	&62.43 \\
    MaPLe~\cite{khattak2023maple} &\blu{79.23}	&\blu{54.33}	&\blu{64.46} \\
    \midrule
    \rowcolor{tabhighlight}
    Ours  &\red{97.73}	&\red{54.83}	&\red{70.25}\\
    \bottomrule
    \end{tabular}
    \end{subtable}
        }
        ~
    \begin{subtable}[t]{.25\textwidth}
    \centering
    \caption{Rock Z dataset}
\includegraphics[width=0.7\linewidth]{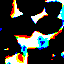}
    \end{subtable}
    \label{table3}
\end{table*}

\begin{table*}[t!]
\small
\tablestyle{6pt}{0}
\addtolength{\tabcolsep}{-6pt}
    \tabstyle{1.0pt}
    \setlength{\tabcolsep}{1pt}
    \caption{Comparison between our method and SOTA methods for fluid image recognition datasets. Our method performs well over the compared methods. We use \red{red} and \blu{blue} to indicate the first and second-best scores.  Except the HM denotes the harmonic mean score, others are the accuracy scores.
    }
    \scalebox{0.7}{
    \begin{subtable}[t]{.25\textwidth}
    \centering
    \caption{\textbf{Average}}
        \begin{tabular}{l cc c}
    \toprule
    & XYZ & UVW & Mean \\
    \midrule
    CLIP~\cite{radford2021learning} &32.94	&33.30	&33.12
 \\
    CoOp~\cite{zhou2022learning} & 35.74	&33.88	&34.81 \\
    CoCoOp~\cite{zhou2022conditional}  &39.41	&34.65	&37.03\\
    MaPLe~\cite{khattak2023maple}  &\blu{41.81}	&\blu{34.75}	&\blu{38.28} \\
    \midrule
    \rowcolor{tabhighlight}
       Ours  &\red{42.57}	&\red{36.37}	&\red{39.47}\\
    \bottomrule
    \end{tabular}

    \end{subtable}
    }
        ~
    \begin{subtable}[t]{.25\textwidth}
    \centering
    \caption{All datasets}
\includegraphics[width=0.72\linewidth]{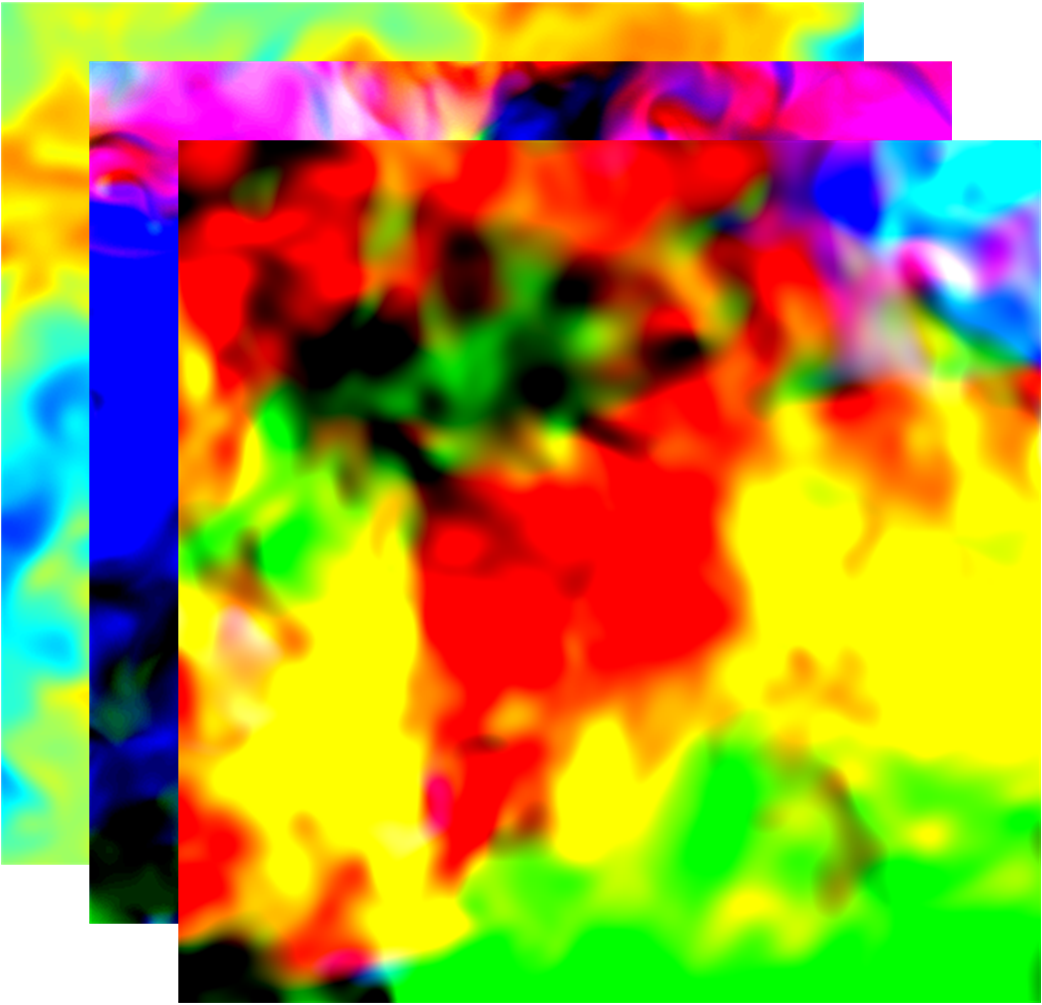}
    \end{subtable}
    ~
\scalebox{0.7}{
    \begin{subtable}[t]{.25\textwidth}
    \centering
    \caption{30000-Step}
    \begin{tabular}{l cc c}
    \toprule
    & XYZ & UVW & Mean \\
    \midrule
    CLIP~\cite{radford2021learning} &32.90	&33.30	&33.10\\
    CoOp~\cite{zhou2022learning} & 35.63	&33.80	&34.72 \\
    CoCoOp~\cite{zhou2022conditional} &39.27	&34.47	&36.87 \\
    MaPLe~\cite{khattak2023maple} &\blu{41.77}	&\blu{34.55}	&\blu{38.16} \\
    \midrule
    \rowcolor{tabhighlight}
     Ours  & \red{42.37}	&\red{36.30}	&\red{39.34}\\
    \bottomrule
    \end{tabular}
    \end{subtable}
    }
            ~
    \begin{subtable}[t]{.25\textwidth}
    \centering
    \caption{30000-Step dataset}
\includegraphics[width=0.7\linewidth]{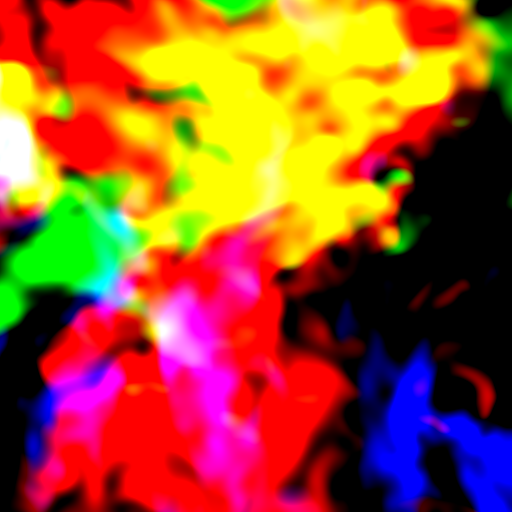}
    \end{subtable}
         ~
        \scalebox{0.7}{
    \begin{subtable}[t]{.25\textwidth}
    \centering
    \caption{40000-Step}
    \begin{tabular}{l cc c}
    \toprule
    & XYZ & UVW & Mean \\
    \midrule
    CLIP~\cite{radford2021learning} &32.95	&33.31	&33.13\\
    CoOp~\cite{zhou2022learning} &35.84	&33.95	&34.90 \\
    CoCoOp~\cite{zhou2022conditional} &39.54	&34.86	&37.20\\
    MaPLe~\cite{khattak2023maple} &\blu{41.82}	&\blu{34.98}	&\blu{38.40} \\
    \midrule
    \rowcolor{tabhighlight}
     Ours  &\red{42.76}	&\red{36.38}	&\red{39.57}
 \\
    \bottomrule
    \end{tabular}
    \end{subtable}
    }
    ~
    \begin{subtable}[t]{.25\textwidth}
    \centering
    \caption{40000-Step dataset}
\includegraphics[width=0.7\linewidth]{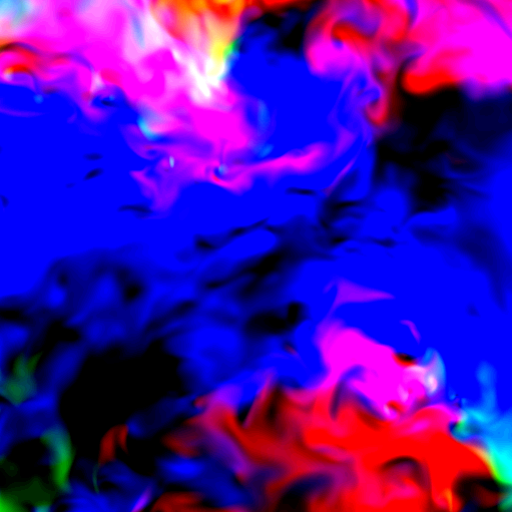}
    \end{subtable}
    ~
        \scalebox{0.7}{
    \begin{subtable}[t]{.25\textwidth}
    \centering
    \caption{50000-step}
    \begin{tabular}{l cc c}
    \toprule
    & XYZ & UVW & Mean \\
    \midrule
    CLIP~\cite{radford2021learning} &32.97	&33.29	&33.13 \\
    CoOp~\cite{zhou2022learning} &35.74	&33.89	&34.82 \\
    CoCoOp~\cite{zhou2022conditional} &39.42	&34.63	&37.03 \\
    MaPLe~\cite{khattak2023maple} &\blu{41.84}	&\blu{34.73}	&\blu{38.29} \\
    \midrule
    \rowcolor{tabhighlight}
    Ours  &\red{42.58}	&\red{36.43}	&\red{39.51}\\
    \bottomrule
    \end{tabular}
    \end{subtable}
        }
        ~
    \begin{subtable}[t]{.25\textwidth}
    \centering
    \caption{50000-step dataset}
\includegraphics[width=0.7\linewidth]{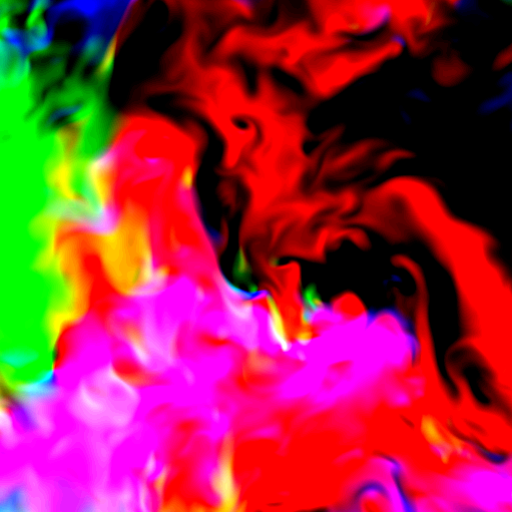}
    \end{subtable}
    \label{table4}
\end{table*}

\section{Detailed Evaluation Results}
\noindent \textbf{Remote Sensing VLM.} The experimental findings within the realm of remote sensing are presented in Table~\ref{table1}. Evidently, by integrating domain knowledge into natural vision-based VLMs, the prompted remote sensing VLM surpasses previously advanced nature vision-based prompt learning methods, yielding an approximate 3\% mean HM performance enhancement. This outcome underscores the efficacy of leveraging domain-specific foundation models to guide the transfer of natural vision-based VLMs into the remote sensing domain. Notably, the most substantial performance enhancements are observed on the RSICD dataset, with an impressive 8.34\% HM gain, while the RSSCN dataset showcases the smallest improvement at 1.39\%. This discrepancy highlights the greater significance of domain-specific foundation models for datasets with lower baselines. Upon closer examination, this trend is discernible across other datasets as well. Of particular interest is the observation that despite ViTAE's greater depth compared to ViT, our method based on ViT achieves superior overall performance. This phenomenon suggests that excessively deep architectures may not be optimal for extracting domain-specific information, as surplus parameters could potentially impede performance.

\noindent \textbf{Medical VLM.} We conduct our proposed domain prompt learning applied to medical datasets, yielding promising experimental results detailed in Table~\ref{table2}. Notably, our method consistently surpasses the performance of previous state-of-the-art nature-vison-based prompt learning methods. Specifically, we observe a remarkable 5.19\% enhancement in mean Harmonic Mean (HM), accompanied by a substantial improvement in base accuracy, escalating from 64.45\% to 80.27\%. Furthermore, our method demonstrates consistent performance gains across diverse medical datasets, indicative of its robustness in transferring Vision-Language Models (VLMs) to various medical domains. Despite these successes, our investigation reveals an intriguing observation: our method does not achieve optimal performance in novel categories. This suggests a limitation in the ability of the utilized medical foundation model to generalize effectively to unseen categories during prompt learning. 

\noindent \textbf{Geology VLM.} Using rock imagery as examples, we validate our proposed domain prompt learning framework within the field of geology, with results summarized in Table~\ref{table3}. Our method consistently outperforms previous prompt learning algorithms by a significant margin across various performance metrics: 12.41\% for base categories, 5.05\% for novel classes, and 8.36\% for the Harmonic Mean (HM). These findings underscore the critical role of introduced domain knowledge in enhancing rock image recognition, affirming the effectiveness of our method in seamlessly transferring natural vision-based Vision-Language Models (VLMs) into the geology domain. Moreover, detailed experiments across three dimensions further highlight the superiority of our proposed generalized domain prompt learning approach. Notably, the most substantial performance enhancement is observed along the Y dimension, with a notable 10.41\% improvement in HM performance.

\noindent \textbf{Fluid Dynamics VLM.} 
Through the application of fluid simulation based on the Navier-Stokes equation, a series of fluid datasets was generated. Flow images obtained at 30,000, 40,000, and 50,000 steps were leveraged for domain prompt learning experiments, with summarized results presented in Table~\ref{table4}. The task of capturing minute-scale structures within the flow field for accurate differentiation presents a formidable challenge. Statistical significance across various directions is minimal, with differentiation achievable solely through discerning fluidity in intricate structures. Despite the modest recognition performance observed across all prompt learning methods, our proposed generalized domain prompt learning approach, fortified by domain-specific foundation models, significantly bolsters the recognition of flow images. Notably, our method exhibits enhancements in average performance, with a 0.76\% refinement in classifying XYZ dimensions, a 1.62\% improvement in UVW dimension classification, and a 1.19\% boost in mean performance. These findings emphasize the efficacy of our approach in advancing flow image recognition, thereby underscoring the pivotal role of domain-specific knowledge in prompt learning tasks.

 \begin{figure*}[ht]
	\begin{center}
		\includegraphics[width=1.0\linewidth]{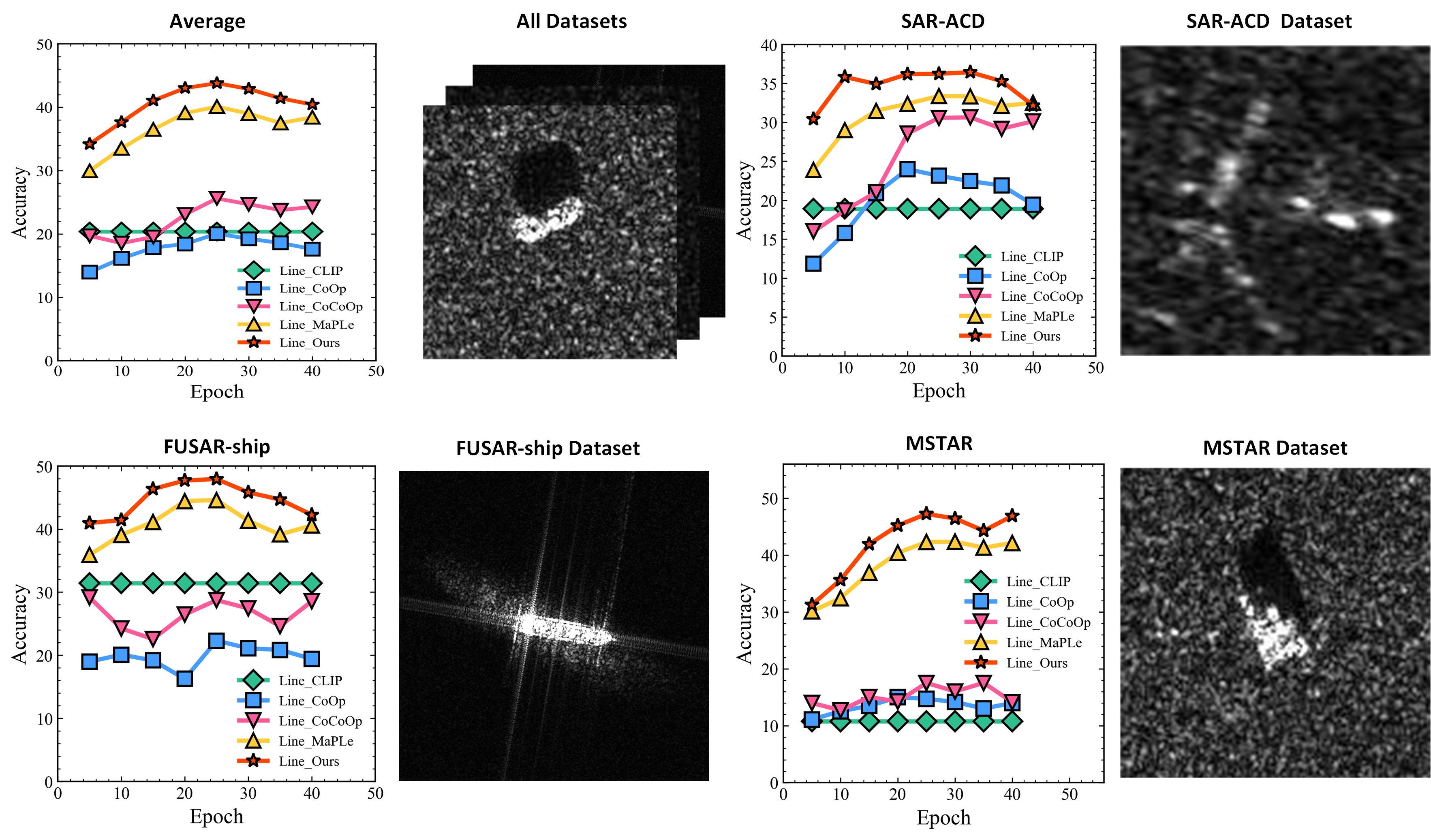}
	\end{center}
	\caption{Comparison between our method and SOTA methods for SAR image recognition datasets. Our method performs well over the compared methods. }
	\label{fig_sar}
\end{figure*}

\begin{table*}[!t]
\small
\centering
\caption{Comparisons with SOTA methods for (a) cross-dataset generalization and (b) single-source multi-target domain generalization with the MLRSNet dataset as the source dataset. The proposed method achieves better performance than the compared methods. \red{Red} and \blu{blue} highlight the first and second best accuracy scores, respectively.}
\begin{subtable}{\linewidth}
\centering
\caption{Cross-dataset generalization}
\scalebox{0.75}{
\begin{tabular}{l c cccccccc}
\toprule
& \textbf{\ \ Source\ \ } & \multicolumn{8}{c}{\textbf{Target}} \\
\cmidrule{2-2} \cmidrule(l){3-10}
& MLRSNet & PatternNet & RSSCN7 & AID & RSICD & UCM & WHURS19 & NWPU & Average\\
\midrule
CoOp~\cite{zhou2022learning} & 72.53 & 66.97 & 69.03 & \red{67.30} & \red{63.50} & 77.57 & \blu{85.47} & 70.43 & 71.60\\
CoCoOp~\cite{zhou2022conditional} & 71.70 & 65.67 & 68.80 & 66.63 & 62.57 & 76.40 & 85.33 & 70.30 & 70.92\\
MaPLe~\cite{khattak2023maple} & \blu{76.83} & \blu{68.53} & \red{71.43} & 65.13 & 59.53 & \blu{79.90} & 85.23 & \blu{72.80} & \blu{72.42}\\
\rowcolor{gray!20} Ours & \red{80.93} & \red{68.70} & \blu{68.97} & \blu{66.93} & \blu{62.97} & \red{80.20} & \red{88.83} & \red{73.27} & \red{73.85}\\
\bottomrule
\end{tabular}
}

\label{table:a}
\end{subtable}

\bigskip

\begin{subtable}{\linewidth}
\centering
\caption{Single-source multi-target domain generalization}
\scalebox{0.75}{
\begin{tabular}{l c cccccccc}
\toprule
& \textbf{\ \ Source\ \ } & \multicolumn{8}{c}{\textbf{Target}} \\
\cmidrule{2-2} \cmidrule(l){3-10}
& MLRSNet & PatternNetv2 & RSSCN7v2 & AIDv2 & RSICDv2 & UCMv2 & WHURS19v2 & NWPUv2 & Average\\
\midrule
CoOp~\cite{zhou2022learning} & 72.53 & 66.97 & 69.07 & \red{67.13} & \red{64.27} & 77.40 & \blu{85.20} & 71.17 & 71.72\\
CoCoOp~\cite{zhou2022conditional} & 71.70 & 65.57 & \blu{69.37} & \red{67.13} & 62.73 & 75.70 & 84.83 & 70.97 & 71.00\\
MaPLe~\cite{khattak2023maple} & \blu{76.83} & \red{68.03} & \red{72.50} & 64.90 & 59.73 & \blu{78.93} & 83.07 & \blu{73.17} & \blu{72.15}\\
\rowcolor{gray!20} Ours & \red{80.93} & \blu{67.90} & 68.01 & \blu{66.87} & \blu{63.93} & \red{79.83} & \red{87.87} & \red{73.83} & \red{73.65}\\
\bottomrule
\end{tabular}
}

\label{table:b}
\end{subtable}

\label{table}
\end{table*}

\noindent \textbf{SAR VLM.} To provide a comprehensive analysis of the performance comparisons of our proposed generalized domain prompt learning approach, we utilize the SAR domain as an illustrative example. We monitor the performance variation across epochs and summarize the experimental results in Figure~\ref{fig_sar}. Interestingly, we observe that while the performance fluctuates, both CoOp and CoCoOp consistently achieve similar or even inferior performance compared to CLIP. In contrast, our method consistently outperforms the compared methods, showcasing robust performance trends across epochs. Notably, we observe an initial performance increase followed by a decline due to overfitting. These phenomena collectively demonstrate the efficacy of our approach in seamlessly transferring vision-based VLMs into the SAR domain, emphasizing its ability to adapt and perform effectively across varying epochs.

\begin{table*}[!t]
\small
    \caption{Ablation Study.}
    \centering
    \begin{subtable}{0.45\textwidth}
	\centering
	\scriptsize
	\footnotesize
	\renewcommand{\tabcolsep}{2.0mm}
	\caption{ The utilization of quaternion network (QN) and fresh low-rank adaptation (LA). w/o denotes without.}
		\begin{tabular}{c|ccc}
			\hline
			\multicolumn{1}{c|}{Methods}  &\multicolumn{1}{c}{Base}  &\multicolumn{1}{c}{Novel}  &\multicolumn{1}{c}{HM} \\
			\hline
      	\multicolumn{1}{c|}{Baseline} &93.93	&56.27	&70.38 \\
          	\multicolumn{1}{c|}{Ours w/o QN, LA} &94.37	&61.98	&74.82\\
   		\multicolumn{1}{c|}{Ours w/o QN} &94.52	&63.53	&75.63\\
            \multicolumn{1}{c|}{Ours w/o LA} &95.18	&65.68	&77.72\\
   		\multicolumn{1}{c|}{Ours}    &\textbf{95.50}	&\textbf{66.95}	&\textbf{78.72}\\
			\hline	
	\end{tabular} 
	\label{table6a}
    \end{subtable}
    \hfill
    \begin{subtable}{0.45\textwidth}
	\centering
	\scriptsize
	\footnotesize
	\renewcommand{\tabcolsep}{2.0mm}
	\caption{ The prompting of vision and language branches. PL: prompting language branch; PV: prompting vision branch.}
		\begin{tabular}{c|ccc}
			\hline
			\multicolumn{1}{c|}{Methods}  &\multicolumn{1}{c}{Base}  &\multicolumn{1}{c}{Novel}  &\multicolumn{1}{c}{HM} \\
			\hline
      	\multicolumn{1}{c|}{Baseline} &93.93	&56.27	&70.38 \\
   		\multicolumn{1}{c|}{PL} &95.31	&63.83	&76.46\\
            \multicolumn{1}{c|}{PV} &95.69	&64.34	&76.94\\
   		\multicolumn{1}{c|}{Ours (PL+PV)}   &\textbf{95.50}	&\textbf{66.95}	&\textbf{78.72}\\
			\hline	
	\end{tabular}
	\label{table6b}
    \end{subtable}
    \label{ablation_study}
\end{table*}

\section{Ablation Study.}
To analyze the impact of various components in our proposed method, we conduct an ablation study using the RSICD dataset from the remote sensing domain. The experimental results are summarized in Table~\ref{ablation_study}. Initially, we investigate the effectiveness of the proposed quaternion network (QN) and the novel low-rank adaptation (LA), presented in Table~\ref{table6a}. Both QN and LA demonstrate performance enhancements, with QN showing more pronounced improvements compared to LA. This suggests that while exploring the domain adaptation potentials of Vision-Language Models (VLMs) is valuable, mining cross-modal relationships is pivotal for effective domain prompt learning. Furthermore, even in the absence of QN and LA, our method achieves commendable performance by directly incorporating domain knowledge through feature augmentation. This underscores the significance of domain knowledge in facilitating the transfer of VLMs from generalized to specific domains. Moreover, we evaluate the impact of prompting vision and language branches individually, as well as jointly, as depicted in (b) of Table~\ref{table6b}. Results reveal that prompting either branch independently yields performance gains owing to enhanced domain-specific vision-language contrastive learning. However, prompting both branches complementarily enables our method to propagate domain-specific information effectively into contrastive learning, resulting in significant performance improvements. In summary, the findings from the ablation study highlight the efficacy of incorporating domain-specific knowledge for transferring VLMs from generalized to specific domains. The quaternion network effectively models orthogonal cross-modal relationships, while the novel low-rank adaptation successfully harnesses the potential of VLMs, ultimately leading to further enhanced performance.

\begin{table*}[h]
\small
    \caption{Network Details.}
    \centering
    \begin{subtable}{0.45\textwidth}
	\centering
	\scriptsize
	\footnotesize
	\renewcommand{\tabcolsep}{2.0mm}
	\caption{ The study of the prompt depth. The best performance is bolded.}
		\begin{tabular}{c|ccc}
			\hline
			\multicolumn{1}{c|}{Depths}  &\multicolumn{1}{c}{Base}  &\multicolumn{1}{c}{Novel}  &\multicolumn{1}{c}{HM} \\
			\hline
      	\multicolumn{1}{c|}{6} &93.83	&62.60	&75.10 \\
          	\multicolumn{1}{c|}{7} &94.23	&63.62	&75.96\\
   		\multicolumn{1}{c|}{8} &95.48	&65.23	&77.51\\
            \multicolumn{1}{c|}{9} &95.50	&\textbf{66.95}	&\textbf{78.72} \\
   		\multicolumn{1}{c|}{10}    &\textbf{95.70}	&65.40	&77.70\\
			\hline	
	\end{tabular} 
	\label{tablena}
    \end{subtable}
    \hfill
    \begin{subtable}{0.45\textwidth}
	\centering
	\scriptsize
	\footnotesize
	\renewcommand{\tabcolsep}{2.0mm}
	\caption{ The study of input patterns of quaternion network, a and b denote the multi-modal inputs, and * denotes the zero tensor .}
		\begin{tabular}{c|ccc}
			\hline
			\multicolumn{1}{c|}{Patterns}  &\multicolumn{1}{c}{Base}  &\multicolumn{1}{c}{Novel}  &\multicolumn{1}{c}{HM} \\
			\hline
      	\multicolumn{1}{c|}{[a,b,*,*]} &95.50	&\textbf{66.95}	&\textbf{78.72} \\
   		\multicolumn{1}{c|}{[*,*,a,b]} &96.03	&63.10	&76.16\\
            \multicolumn{1}{c|}{[a,*,*,b]} &95.83	&62.90	&75.95\\
   		\multicolumn{1}{c|}{[a,*,b,*]}   &95.57	&62.53	&75.60\\
        	\multicolumn{1}{c|}{[*,a,*,b]}   &\textbf{95.90} 	&63.23	&76.21\\
             \multicolumn{1}{c|}{[*,a,b,*]}   &95.80	&63.80	&76.59\\
			\hline	
	\end{tabular}
	\label{tablenb}
    \end{subtable}
    \label{Network Details}
\end{table*}

\section{Network Details}

In our training process, to ensure fair performance comparison, all input images are resized to a uniform shape of 224 × 224. We utilize a learning rate of 3.5e-3 with the SGD optimizer, while setting the batch size to 4 during training and 24 during testing. The training process continues for either 15 or 10 epochs, depending on convergence, and the length of learnable contexts is restricted to 2 words.

To determine the appropriate prompting depth for generalized domain prompt learning, experiments are conducted on the RSICD dataset, and the results are presented in Table~\ref{tablena}. The performance improvements for base categories correspond to increasing prompt depth; however, when the depth reaches 9, novel categories achieve peak performance, indicating a potential overfitting issue with deeper prompts. Therefore, the prompting depth is set to 9 for our experiments.

Additionally, considering various input patterns in the quaternion network, experimental results are shown in Table~\ref{tablenb}. Among the input patterns, [*] denotes a zero tensor, [a] and [b] denote multi-modal inputs. Notably, the [*, a,*,b] pattern exhibits the highest base accuracy but relatively low novel accuracy, indicating potential overfitting and weak robustness. Conversely, the [a,b,*,*] pattern demonstrates the best novel and HM accuracy, hence chosen as the experimental pattern for further experiments.

Scrutinizing the results further, despite variations in accuracy with different input patterns, the model consistently outperforms the baseline without a quaternion network, underscoring the significance of mining orthogonality between multi-modal features.

\end{document}